%% file: main.tex
\documentclass[10pt,twocolumn,letterpaper]{article}

\usepackage[pagenumbers]{cvpr} 

\input{preamble}

\definecolor{cvprblue}{rgb}{0.21,0.49,0.74}
\usepackage[pagebackref,breaklinks,colorlinks,allcolors=cvprblue]{hyperref}

\title{MUOT-3M --- A 3 Million Frame Multimodal Underwater Benchmark \\ and the MUTrack Tracking Method }

\author{
Ahsan Baidar Bakht$^{1}$ \quad
Mohamad Alansari$^{1}$ \quad
Muhayy Ud Din$^{1}$ \quad
Muzammal Naseer$^{1}$ \\
Sajid Javed$^{1}$ \quad
Irfan Hussain$^{1}$ \quad
Jiri Matas$^{2}$ \quad
Arif Mahmood$^{3}$ \\
\\
$^{1}$Khalifa University, Abu Dhabi, UAE \\
$^{2}$Czech Technical University, Prague, Czech Republic \\
$^{3}$Information Technology University, Lahore, Pakistan \\
}

\begin{document}

\maketitle

\begin{teaserfigure}
  \centering
  \includegraphics[width=\textwidth]{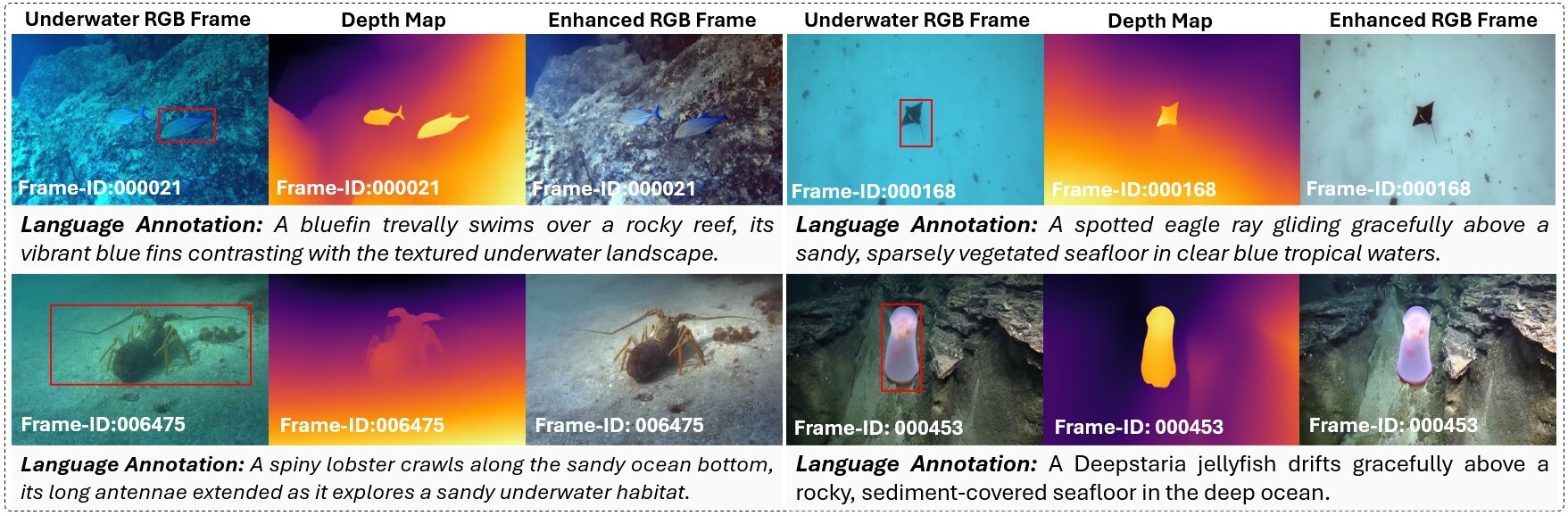}
  \vspace{-2em}
  \captionof{figure}{\textbf{MUOT-3M dataset sample images}. The language annotations are validated by an expert marine biologist.}
  \label{fig:teaser}
\end{teaserfigure}

\begin{figure*}[t!]
    \begin{minipage}{0.73\linewidth}
        \centering \small
        \includegraphics[width=\linewidth]{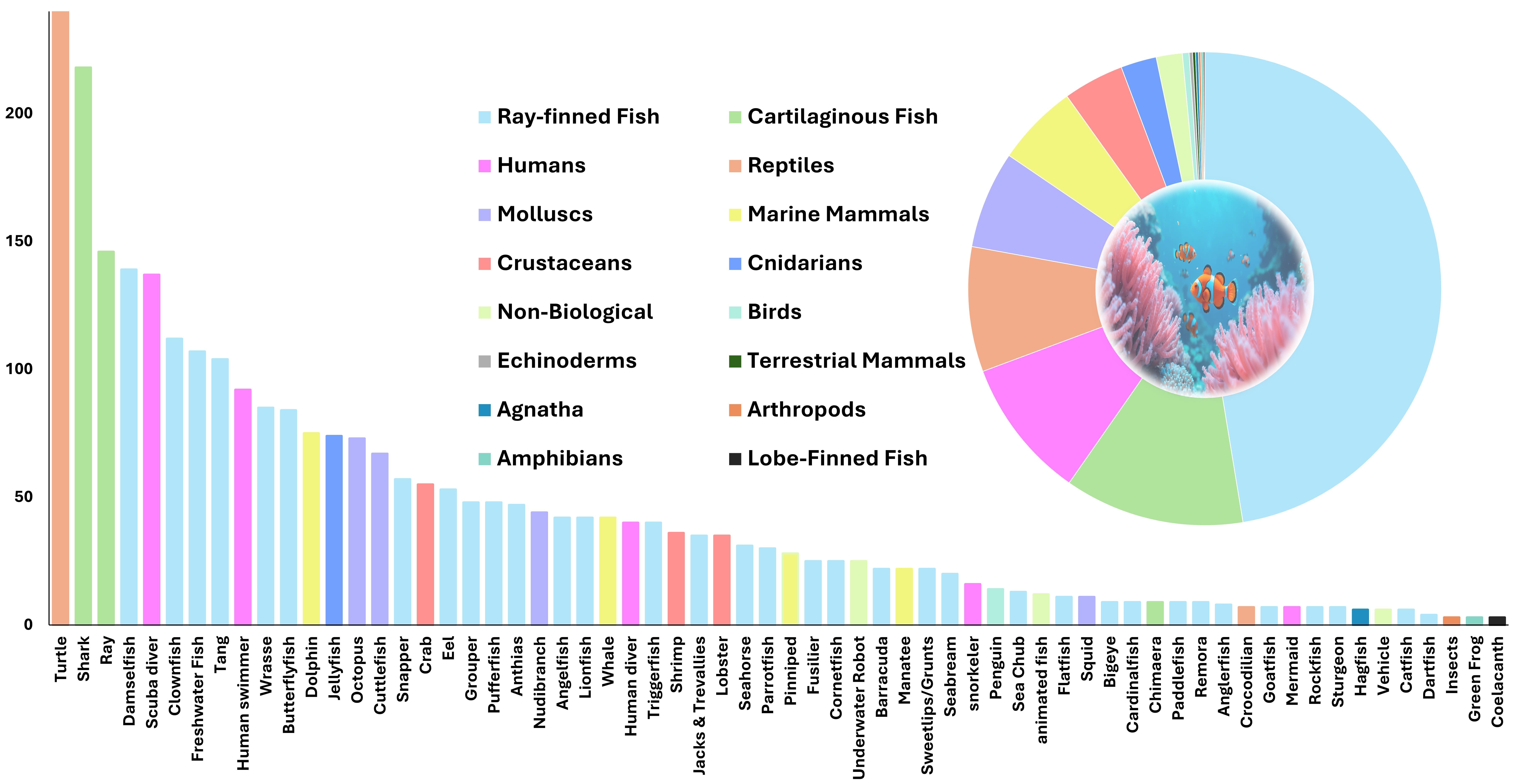}
    \end{minipage}
    \hfill
    \begin{minipage}{0.29\linewidth}
            \caption{\textbf{MUOT-3M dataset} diversity in terms of 16 Phylum categories, 124 families, and 677 fine-grained classes. 16 Phylum categories with corresponding representative families are shown. \textit{The distribution and labels of all classes are validated by the expert marine biologist}.
    Non-marine species categories in MUOT-3M, i.e., human-related (diver, scuba) and non-biological (robot, ROVs), are not shown.}
        \label{fig_diversity}
    \end{minipage}
    \vspace{-2em}
\end{figure*}

\input{sec/0_abstract}

\vspace{-1em}
\input{sec/1_intro}

\input{sec/Relatedwork}

\input{sec/MUOT}

\input{sec/MAquaTracker}

\input{sec/results}
\input{sec/conclusion}

{
    \small
    \bibliographystyle{IEEEtran}
    \bibliography{main}
}

\clearpage
\appendix
\section*{Supplementary Material}
\addcontentsline{toc}{section}{Supplementary Material}

\input{supplementary}

\end{document}

%% file: preamble.tex
\usepackage{xcolor}

\usepackage{pifont}     
\usepackage{svg}        
\usepackage{comment}    
\usepackage{multicol}   
\usepackage{multirow}   
\usepackage{placeins} 
\usepackage{cuted}
\newenvironment{teaserfigure}{\begin{strip}\centering}{\end{strip}}
\usepackage[font=small,labelfont=bf]{caption}

\usepackage{amsmath,amssymb}
\usepackage{booktabs}    
\usepackage{graphicx}    
\usepackage{microtype}   



%% file: sec/0_abstract.tex
\begin{abstract}
Underwater Object Tracking (UOT) is crucial for efficient marine robotics, large-scale ecological monitoring, and ocean exploration; however, progress has been hindered by the scarcity of large, multimodal, and diverse datasets. Existing benchmarks remain small and RGB-only, limiting robustness under severe color distortion, turbidity, and low-visibility conditions.We introduce \textbf{MUOT-3M}, the first pseudo-multimodal UOT benchmark comprising 3 million frames from 3,030 videos (27.8h) annotated with 32 tracking attributes, 677 fine-grained classes, and synchronized RGB, estimated enhanced RGB, estimated depth, and language modalities validated by a marine biologist. Building upon \textbf{MUOT-3M}, we propose \textbf{MUTrack}, a SAM-based multimodal-to-unimodal tracker featuring visual–geometric alignment, vision-language fusion, and four-level knowledge distillation that transfers multimodal knowledge into a unimodal student model.Extensive evaluations across five UOT benchmarks demonstrate that MUTrack achieves up to 8.40$\%$ higher AUC and 7.80$\%$ higher precision than the strongest SOTA baselines while running at 24 FPS.\textbf{MUOT-3M} and \textbf{MUTrack} establish a new foundation for scalable, multimodally trained yet practically deployable underwater tracking.
\footnote{Project Link: \url{https://github.com/AhsanBaidar/MUOT-3M_Dataset}}
\end{abstract}
\vspace{-1em}

%% file: sec/1_intro.tex
\section{Introduction}
\label{sec:intro}
Visual Object Tracking (VOT) in terrestrial environments aims to estimate the trajectory of a target object given its position in the first frame \cite{javed2022visual, marvasti2021deep, chen2022visual, abdelaziz2025beyond, din2024benchmarking}.
Underwater Object Tracking (UOT) extends VOT to the submerged domain, where optical degradation fundamentally reshapes tracking dynamics \cite{gonzalez2023survey, alawode2022utb180, zhang2024webuot, kezebou2019underwater,din2025maritime}.
UOT is important for marine robotics \cite{zereik2018challenges, akram2024enhancing}, autonomous exploration \cite{cimurs2021goal, ELMEZAIN2026119384,KHAN2025126820}, and search and rescue operations \cite{mansor2021autonomous,ahmed2023vision}, yet it remains underexplored within the VOT community \cite{kristan2015visual,kristan2019seventh}.

\begin{figure}[t!]
\begin{minipage}{0.70\linewidth}
   \centering \small
    \includegraphics[width=\columnwidth]{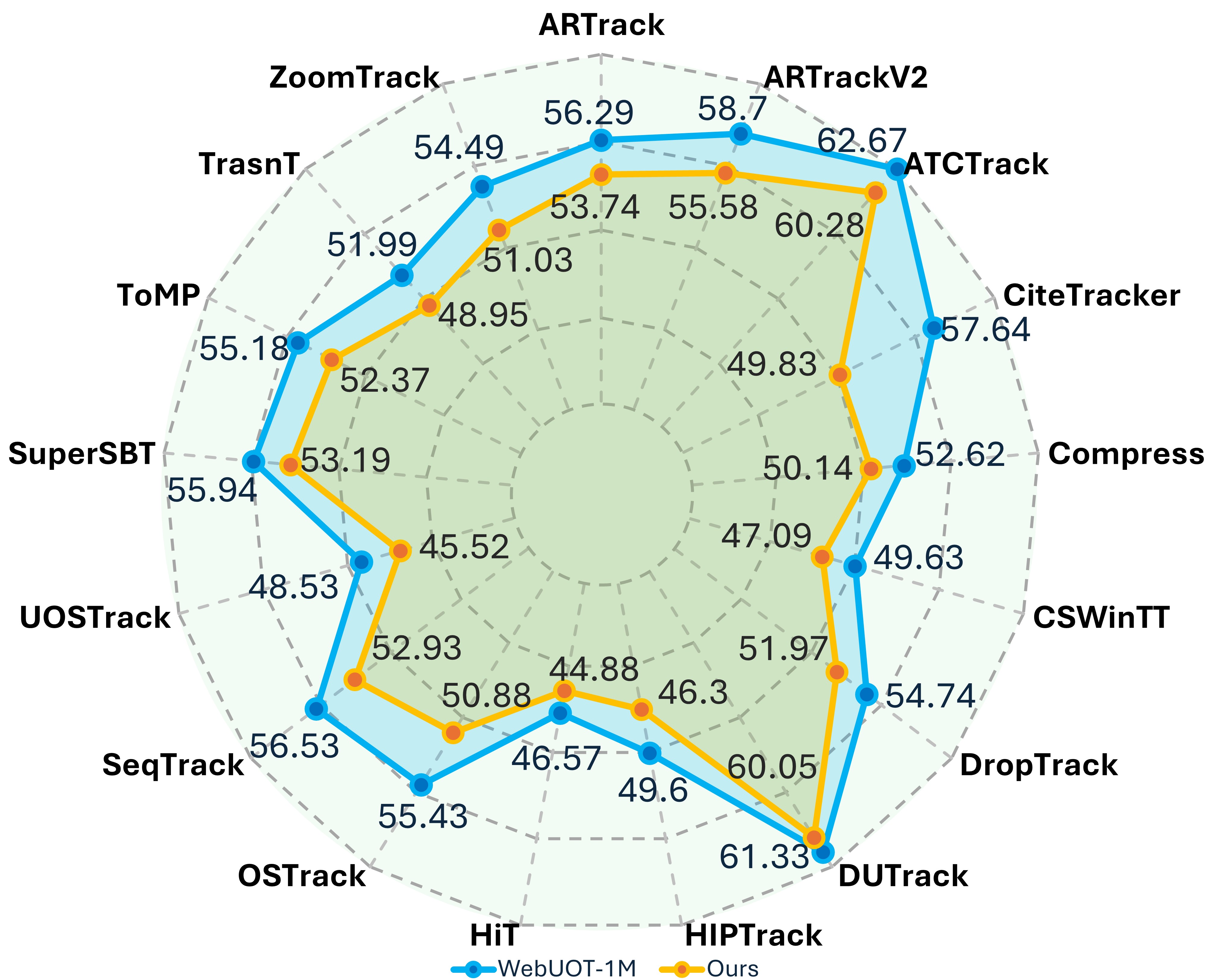}
        \end{minipage}
    \hfill
    \begin{minipage}{0.29\linewidth}
\caption{Performance degradation of SOTA trackers on WebUOT-1M \cite{zhang2024webuot} and MUOT-3M.}
    \label{fig:quadratic_chart}
      \end{minipage}
\vspace{-2em}
\end{figure}

Significant progress has been achieved through deep learning trackers \cite{fan2019siamese, li2018high}, supported by terrestrial environments' large-scale datasets, including LaSOT \cite{fan2019lasot, fan2021lasot} and TrackingNet \cite{muller2018trackingnet}.
Datasets and trackers tuned to terrestrial environment challenges fail under the severe color distortion, turbidity, and low visibility of underwater scenes (Fig. \ref{fig:quadratic_chart}) \cite{gonzalez2023survey, zhang2024webuot, nizami2020natural, nizami2020no}.
\textit{This performance gap underscores the need for an underwater-specific dataset and tracking framework for VOT in degraded environments.}

Underwater imagery presents,
unlike terrestrial environments,
inherent challenges, including \textit{light scattering}, \textit{color absorption, nonuniform illumination}, and \textit{dynamic water patterns}, which severely degrade \textit{contrast} and \textit{distort object appearance} \cite{islam2024computer, gonzalez2023survey, gracias2017application, fu2023rethinking,rehman2024iqa, zhang2024webuot, nasir2025self, REHMAN2026114}.
Existing UOT benchmarks remain small and RGB-only, which limits their robustness under severe color distortion, turbidity, and low-visibility conditions \cite{alawode2022utb180, zhang2024webuot, REHMAN2024580}.

Despite UOT's significant importance, progress has been limited due to \textit{lack of large-scale, diverse, and multimodal datasets} that comprehensively capture underwater visual conditions.
Several UOT benchmarks have recently been introduced to address these challenges \cite{alawode2023improving, zhang2024webuot}.
Prominent datasets are UOT32 \cite{kezebou2019underwater}, UOT100 \cite{panetta2021comprehensive}, UTB180 \cite{alawode2022utb180}, VMAT \cite{cai2023semi}, UVOT400 \cite{alawode2023improving}, and the large-scale WebUOT-1M \cite{zhang2024webuot}, contributing valuable initial insights but offering limited visual and behavioral diversity.
For instance, benchmarks like UTB180 \cite{alawode2022utb180} and UVOT400 \cite{alawode2023improving} expanded the number of categories and tracking attributes, while WebUOT-1M further increased scale to over a million frames and introduced language prompts for Vision–Language (VL) tracking \cite{zhang2024webuot}.
\textit{Nevertheless, all existing benchmarks remain RGB-only, lacking critical depth, enhanced, and semantic modalities required for comprehensive UOT.}
Even the an large-scale WebUOT-1M benchmark still exhibits limited scene diversity, capturing only a subset of visibility conditions and object categories \cite{zhang2024webuot}.
As a result, current benchmarks fail to represent the geometric, photometric, and semantic complexity of real underwater environments, limiting progress toward multimodal and domain-specific UOT frameworks.

\textit{In the current work, we address this gap by introducing a Multimodal Underwater Object Tracking dataset containing 3 Million frames (\textbf{MUOT-3M}).}
We also proposed a novel \textit{Multimodal UOT framework (\textbf{MUTrack})}, designed for underwater environments, \textit{leveraging multimodal training and unimodal inference}.

\begin{figure}[t!]
\centering
\includegraphics[width=\columnwidth]{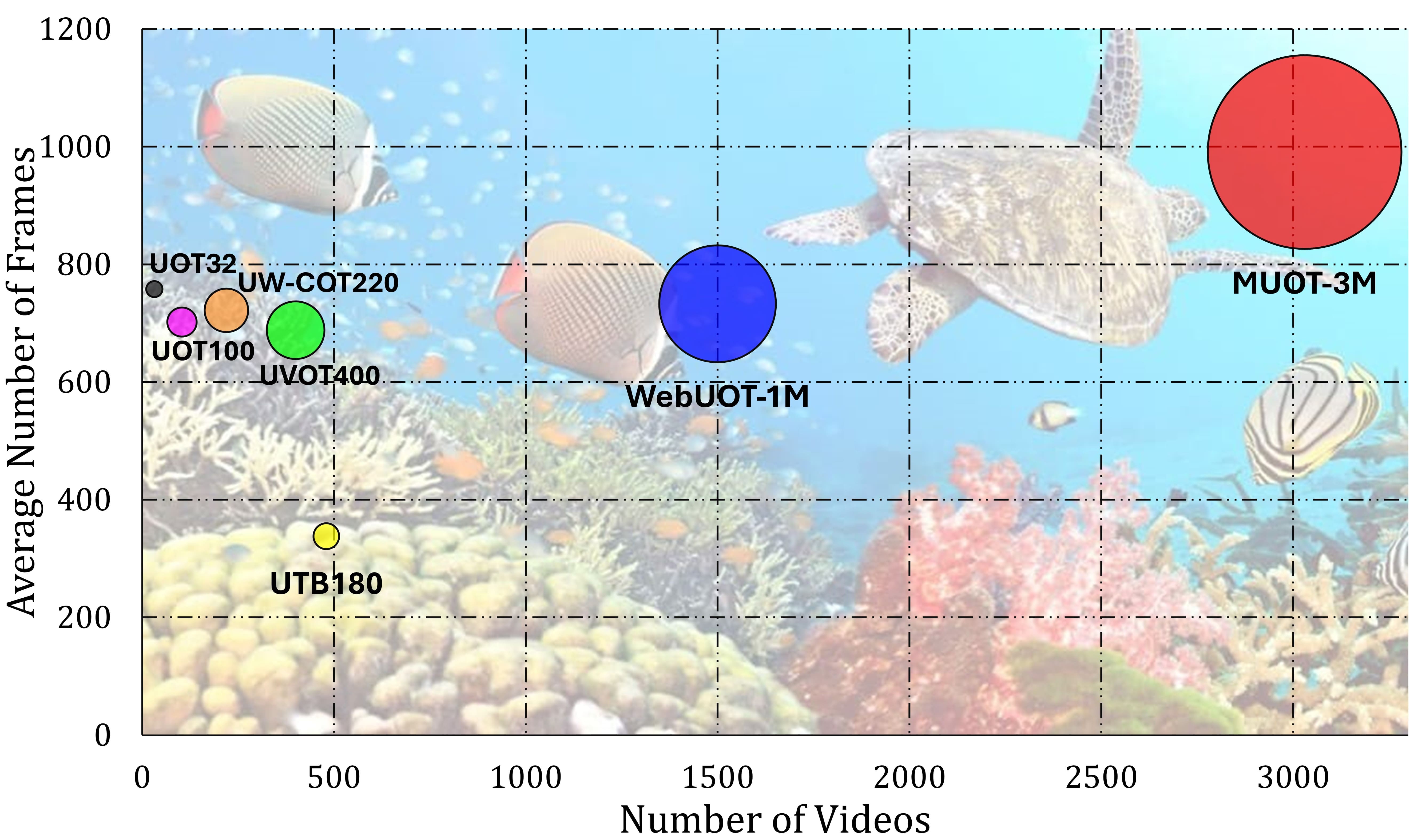}
\caption{\textbf{MUOT-3M} is much larger than existing UOT datasets.}
\label{fig_sota_trackers}
\end{figure}

\textbf{MUOT-3M} is \textit{three times larger} than the current largest UOT dataset, WebUOT-1M (Fig. \ref{fig_sota_trackers}), and offers multiple modalities, increased categorical and environmental diversity (Table \ref{table1}).
It comprises \textit{3,030 underwater videos (27.8 hours)} with \textit{RGB frames, estimated enhanced RGB frames and depth maps with linguistic modalities}, enabling visual–geometric and semantic alignments (Fig. \ref{fig:teaser}).
\textbf{MUOT-3M} is densely annotated with high-quality bounding boxes, language descriptions, and 32 tracking attributes encompassing conditions such as low visibility, color attenuation, motion blur, and dynamic illumination.
\textbf{MUOT-3M} includes \textit{16 Phylum categories, 124 Families, 677 fine-grained species} with reference to WordNet \cite{miller1995wordnet}, representing a broad spectrum of marine species and objects captured under varying optical conditions, depths, and camera motions (Fig. \ref{fig_diversity}).
To ensure diversity and quality, videos are sourced from multiple online platforms.
Using a filtering process, only high-quality sequences are included in the MUOT-3M dataset, which are then annotated by a 16-member expert team following a rigorous verification protocol.
By combining scale, modality, and semantic richness, \textit{MUOT-3M provides a comprehensive foundation for exploring  cross-modal representation learning for UOT.}

Building on MUOT-3M, we present \textbf{MUTrack}, a SAM-based \cite{kirillov2023segment} multimodal-to-unimodal tracker designed to learn cross-modal representations that generalize across degraded underwater environments. 
MUTrack pipeline consists of three stages, including \textit{visual-geometric and visual-language alignments}, \textit{finetuning of the SAM-driven multimodal teacher tracker}, and the \textit{unimodal (RGB-only) SAM-driven student tracker} that distills multimodal knowledge from the teacher tracker. 
In the first stage, dual encoders are pre-trained on enhanced RGB frames, depth maps, and language representation to enforce visual-geometric and visual-language alignments through a combination of feature-level $\ell_{1}$ and contrastive pre-training objectives, 
This stage enables the model to jointly learn \textit{color information, structural geometry, and illumination variations}, thereby building a modality-invariant feature space that is robust to underwater distortions.
In the second stage, the SAM-based teacher tracker is finetuned using \textit{multimodal feature representations} for underwater target segmentation.
In the third stage, a \textit{unimodal underwater RGB-only student tracker} is pre-trained using multi-level Knowledge Distillation (KD) losses to mimic the teacher tracker. 
Specifically, four complementary KD objectives, including teacher max-pooled visual-geometric features, spatiotemporal attention, VL adapter embeddings, and segmentation masks are proposed to transfer multi-modal knowledge from teacher to student.
\textit{MUTrack leverages enhanced RGB, depth, and language cues only during training, while at inference, it operates on underwater RGB frames, reflecting real-world conditions where auxiliary modalities are unavailable. }

Extensive experiments are performed using protocols suggested by LaSOT \cite{fan2019lasot, fan2021lasot}.
Results are compared with 20 SOTA trackers, pretrained on terrestrial environment datasets.
Moreover, we evaluate five representative SOTA trackers using the training and testing splits of MUOT-3M datasets. 
\textit{Our experiments demonstrate that pre-training on MUOT-3M significantly enhances robustness and generalization, confirming the necessity and value of large-scale multi-modal datasets for developing UOT pipeline.\\ Our main contributions are:}

\begin{enumerate}
\item The \textbf{\textit{MUOT-3M}} dataset, a 3-million-frame multimodal benchmark consisting of 3,030 videos for scale, diversity, and modality for UOT.
\item \textbf{\textit{MUTrack}}, a SAM-based multimodal-to-unimodal tracking framework that bridges multimodal pretraining and unimodal inference.
\item A \textbf{\textit{Teacher tracker}} that leverages multimodal representations and input them into the SAM for target segmentation learning degradation-invariant and semantically grounded UOT.
\item A \textbf{\textit{Unimodal student tracker}} that leverages underwater RGB-only frames and mimic multimodal knowledge from teacher using four complementary knowledge distillation objectives.
\end{enumerate}


\begin{table*}[h!]
\centering
\setlength{\tabcolsep}{2pt}
\renewcommand{\arraystretch}{1.0}
\begin{tabular}{l c r r r r r r r r c c c c}
\toprule
Dataset & Year & Videos & Classes & Attr. & \begin{tabular}[c]{@{}c@{}}Min\\ Frames\end{tabular} & \begin{tabular}[c]{@{}c@{}}Mean\\ Frames\end{tabular} & \begin{tabular}[c]{@{}c@{}}Max\\ Frames\end{tabular} & \begin{tabular}[c]{@{}c@{}}Total\\ Frames\end{tabular} & \begin{tabular}[c]{@{}c@{}}Total\\ Duration\end{tabular} & \begin{tabular}[c]{@{}c@{}}Depth\\ Maps\end{tabular} & \begin{tabular}[c]{@{}c@{}}Target\\ Description\end{tabular} & Train & Test \\
\midrule
UOT32 & 2019 & 32 & - & - & 283 & 758 & 1,573 & 24K & 0.27 h & \ding{55} & \ding{55} & \ding{55} & \ding{51} \\
UOT100 & 2022 & 104 & - & 3 & 264 & 702 & 1,764 & 74K & 0.68 h & \ding{55} & \ding{55} & \ding{55} & \ding{51} \\
UTB180 & 2022 & 180 & - & 10 & 40 & 338 & 1,226 & 58K & 0.53 h & \ding{55} & \ding{55} & \ding{51} & \ding{51} \\
VMAT & 2023 & 33 & 17 & 13 & 438 & 2,242 & 5,550 & 74K & 0.68 h & \ding{55} & \ding{55} & \ding{55} & \ding{51} \\
UVOT400 & 2023 & 400 & 50 & 17 & 40 & 688 & 3,273 & 275K & 2.6 h & \ding{55} & \ding{55} & \ding{51} & \ding{51} \\
WebUOT-1M & 2024 & 1,500 & 408 & 23 & 49 & 733 & 9,985 & 1.1M & 10.5 h & \ding{55} & \ding{51} & \ding{51} & \ding{51} \\
UW-COT220 & 2025 & 220 & 96 & - & 10 & 722 & 7,448 & 159K & 1.5 h & \ding{55} & \ding{51} & \ding{51} & \ding{51} \\
MUOT-3M & 2025 & 3,030 & 677 & 32 & 102 & 991 & 16,322 & 3.01M & 27.8 h & \ding{51} & \ding{51} & \ding{51} & \ding{51} \\
\bottomrule
\end{tabular}
\caption{Underwater object tracking datasets. For the attributes provided with a specific dataset, see supplementary, Fig. S7.}
\label{table1}
\end{table*}

%% file: sec/Relatedwork.tex
\section{Related Work}
\label{sec:relatedwork}
\textbf{1. Terrestrial Environment VOT Datasets:} The rapid progress in terrestrial environment-specific VOT has been largely influenced by the availability of large-scale benchmarks \cite{kristan2024second,AFZAL2025102950, kristan2020eighth, kristan2019seventh,bakht2025mvtd, kristan2015visual}.
Prominent datasets are OTB100 \cite{wu2013online}, UAV123 \cite{benchmark2016benchmark}, OxUvA \cite{Valmadre_2018_ECCV}, NUS-PRO \cite{li2015nus}, TrackingNet \cite{muller2018trackingnet}, LaSOT \cite{fan2019lasot, fan2021lasot}, and GOT-10K \cite{huang2019got} etc., to name a few.
These benchmarks have played an important role in standardizing evaluation protocols, diversifying object categories, and enabling data-driven training at scale.
However, existing VOT datasets remain inherently restricted to terrestrial environments and fail to encompass the visual degradations such as optical distortions, low-contrast imagery, and dynamic visibility fluctuations, natural characteristics of the marine environments \cite{alawode2023improving, zhang2024webuot}.
\textit{This limitation underscores the need for new benchmarks that extend beyond terrestrial visibility, paving the way for robust VOT in more challenging domains such as the underwater environment.}



\noindent \textbf{2. UOT Datasets:} UOT remains a comparatively underexplored research area compared to terrestrial VOT \cite{gonzalez2023survey, alawode2022utb180, zhang2024webuot, kezebou2019underwater}.
It is primarily due to the difficulty of collecting and annotating high-quality underwater video data \cite{gonzalez2023survey}.
In the literature, several generic UOT datasets have been proposed including UOT32 \cite{kezebou2019underwater}, UOT100 \cite{panetta2021comprehensive}, UTB180 \cite{alawode2022utb180}, UVOT400 \cite{alawode2023improving}, VMAT \cite{cai2023semi}, and WebUOT-1M \cite{zhang2024webuot}.
VMAT focused on marine animal tracking with temporal annotations, while WebUOT-1M significantly expanded the scale to over one million frames across 1,500 video clips and 408 object categories, incorporating language descriptions to facilitate VL tracking.
Despite this progress, existing UOT datasets remain RGB-only and capture only a subset of underwater visibility conditions and object types, limiting their ability to represent the geometric, photometric, and semantic variability of real underwater scenes \cite{alawode2022utb180, alawode2023improving, zhang2024webuot}.
As a result, current benchmarks are insufficient for training and evaluating multi-modal or domain-specific trackers capable of operating across the full spectrum of underwater challenges.
\textit{In this work, we fill this gap by proposing a new  MUOT-3M dataset which is not only large and diverse but also multi-modal by design.}

\noindent \textbf{3. VOT Paradigms:} Over the past decade, several VOT paradigms have been emerged, reflecting the rapid evolution of representation learning \cite{javed2022visual, abdelaziz2025beyond, chen2022visual}.
These include Discriminative Correlation Filter (DCF)–based trackers \cite{henriques2014high}, Siamese-based trackers \cite{bertinetto2016fully}, Vision Transformer (ViT)–based trackers \cite{chen2021transformer, lin2022swintrack}, and Vision–Language (VL)–based trackers \cite{zheng2023toward,guo2022divert, li2023citetracker}.
Representative DCF trackers are SRDCF \cite{danelljan2015learning}, DiMP \cite{bhat2019learning}, and PrDiMP \cite{danelljan2020probabilistic}.
Siamese trackers are SiamFC \cite{bertinetto2016fully}, SiamRPN \cite{li2018high}, and SiamMask \cite{wang2019fast}.
ViT-based trackers are TransT \cite{chen2021transformer}, STARTK \cite{yan2021learning}, KeepTrack \cite{mayer2021learning}, ToMP \cite{ToMP}, and MixFormer \cite{cui2022mixformer} etc.
Similarly, the representative VL trackers are ATCTrack \cite{ATCTrack}, DUTrack \cite{DUTrack}, CiteTracker \cite{citetracker}, and JointNL \cite{JointNL}.
Each paradigm introduces a distinct modeling approach,  progressively advancing the VOT field from handcrafted feature matching toward data-driven, context-aware, and semantically guided tracking \cite{kristan2024second, kristan2020eighth, kristan2019seventh, kristan2015visual}.

Recently, SAM tracking paradigms have gained attention for their ability to leverage large-scale segmentation pretraining for zero-shot or few-shot tracking \cite{ding2025sam2long, yang2024samurai, videnovic2025distractor}.
These models introduce a generalized, prompt-driven framework that adapt to diverse tracking tasks without domain-specific fine-tuning, bridging the gap between segmentation and tracking paradigms.
The aforementioned VOT paradigms have progressively enhanced efficiency, accuracy, and contextual understanding, yet they remain limited by their dependence on single-modality RGB cues learned from terrestrial domains.
Though several RGB-D trackers have been proposed based on the existing tracking paradigms \cite{yan2021depthtrack, yang2022rgbd}; yet these trackers are underexplored in the submerged domain.
\textit{Our MUTrack fills this gap, demonstrating that multi-modal pretraining can be used for uni-modal inference showing accuracy and robustness in underwater environment.}

%% file: sec/MUOT.tex
\section{Proposed MUOT-3M Dataset}
\label{sec:MUOT}
\label{-1em}

\textbf{Dataset Statistical Details:} Table \ref{table1} compares the main statistics of proposed MUOT-3M dataset with existing UOT benchmarks.
Particularly, \textbf{MUOT-3M} contains \textbf{3,030} videos with \textbf{3 million} frames spanning \textbf{27.8 hours}, approximately three times larger than the existing dataset \cite{zhang2024webuot}.
Each underwater RGB frame corresponds to the estimated enhanced RGB frame, estimated depth map, and language description, offering a unified multimodal representation that jointly captures photometric, geometric, and semantic cues.
The dataset spans frame resolutions ranging from $\mathbf{720 \times 1280}$ to $\mathbf{2160 \times 3840}$ pixels captured at \textbf{30} fps, reflecting the variations in underwater scenes.
Approximately \textbf{37}$\%$ of sequences exhibit low-visibility or backscatter, \textbf{42}$\%$ moderate clarity, and \textbf{21}$\%$ high-visibility conditions.
\textit{Such balanced representation ensures that trackers are not biased toward clear-water imagery, a limitation common in previous UOT benchmarks.}

MUOT-3M contains \textit{\textbf{16} Phylum categories, \textbf{124} families, and \textbf{677} fine-grained target classes} (Fig. \ref{fig_diversity}), with an average sequence length of \textbf{991} frames.
It also contains \textbf{32} tracking attributes covering both standard VOT and underwater-specific attributes.
MUOT-3M further includes absent-frame labels, enabling the evaluation of long-term re-identification.
\textit{MUOT-3M is fully open-source with train/test partitions, providing a scalable platform for multimodal training and unimodal inference in UOT.}
\\
\textbf{MUOT-3M Collection:} Videos are sourced from a broad range of online platforms and media archives, including \textit{YouTube, BiliBili, Netflix, National Geographic, Pixabay, and social media outlets such as Facebook, Instagram, and TikTok.}
This web-scale strategy enabled access to footage recorded by professional filmmakers, marine scientists, and recreational divers across diverse geographical regions.
To ensure comprehensive environmental coverage, we performed targeted keyword searches on these platforms using terms such as \textit{underwater world, marine documentary, deep oceans, Great Barrier Reef, aquatic scenes, and coral reefs.}
\textit{By leveraging these heterogeneous sources, MUOT-3M covers diverse marine environments—coastal shallows, pelagic waters, estuaries, lakes, and rivers with varying conditions of illumination, turbidity, and color attenuation}.

\noindent \textbf{MUOT-3M Filtering Process:} After collection, all videos underwent a comprehensive manual curation to ensure visual quality, temporal coherence, and ecological diversity.
From over 20K videos, 3,030 sequences are retained following expert reviews by marine biologists and 16 graduate students in computer vision and marine ecology.
Each sequence is examined frame-by-frame to confirm continuous target visibility for at least 100 frames, removing clips with static scenes, surface footage, or non-underwater content.
Only single-shot sequences are preserved.
The reviewers balance visibility levels and target categories to capture the full spectrum of underwater variability.
\textit{This expert-driven filtering process ensures MUOT-3M comprises large-scale, high-quality, diverse data for multimodal UOT research.}

\noindent \textbf{Multimodal Representations:} Each sequence in MUOT-3M provides five modalities, including underwater RGB, enhanced RGB, depth maps, segmentation, and language description.
\textit{Depth maps are estimated using MiDaS \cite{ranftl2020towards}, producing dense, geometrically consistent predictions from monocular underwater imagery.
Enhanced RGB frames are generated using UTransformer and Mula-GAN \cite{peng2023u, bakht2024mula}, enhancing color fidelity and contrast while mitigating backscatter, haze, and color attenuation.
Segmentation masks are generated using the SAM model, followed by manual inspection.
For the language modality, representative frames are captioned by GPT4 \cite{hurst2024gpt}, verified by marine biologists.}

\noindent \textbf{Annotations and Attributes:} Bounding box annotations are performed using RGB frame by the same team.
We adopted a semi-supervised approach using the SOTA DAM4SAM tracker \cite{Videnovic_2025_CVPR} to generate initial bounding boxes, which are then manually verified and refined frame-by-frame to correct drift and occlusion errors.
\textit{A final validation round by the expert team ensured annotation precision and consistency.}
\textit{MUOT-3M dataset is annotated with \textbf{32} tracking attributes covering both underwater-specific and standard terrestrial challenges.}
Among these, \textbf{15} underwater attributes capture inherent marine complexities such as swarm distractors, camouflage, artificial objects, varying underwater visibility (high, medium, low), water color (green, blue, yellow), camera position (submerged, above water), bubbles, turbidity, reflections, and transparency.
The remaining \textbf{17} attributes represent conventional VOT factors, including occlusion, motion blur, and scale variation, etc.
The target objects are annotated with \textbf{16} Phylum categories, \textbf{124} families, and \textbf{677} fine-grained classes (Fig. \ref{fig_diversity}).
This hierarchical taxonomy spans both biological and non-biological entities, covering marine fauna such as ray-finned fish, cartilaginous fish, reptiles, molluscs, crustaceans, marine mammals, and amphibians, as well as inanimate targets including divers, robots, and vehicles.
\textit{As shown in Fig. \ref{fig_diversity}, dominant phylum categories include ray-finned fish, cartilaginous fish, and reptiles, reflecting their prevalence in the underwater environment.}

%% file: sec/MAquaTracker.tex
\section{Proposed MUTrack}
\label{sec:tracker}

\begin{figure*}[t!]
     \begin{minipage}{0.73\linewidth}
   \centering \small
    \includegraphics[width=\textwidth]{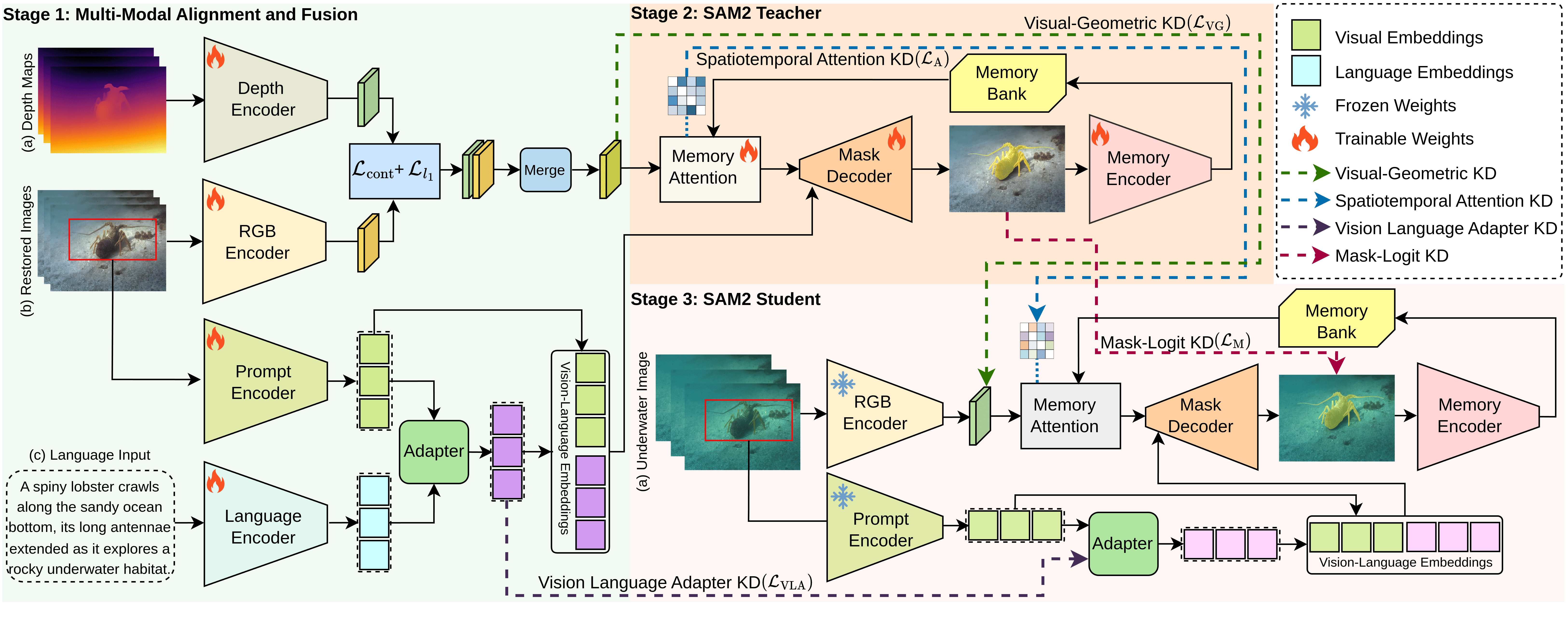}
     \end{minipage}
    \hfill
    \begin{minipage}{0.29\linewidth}
    \caption{\textbf{MUTrack:} Schematic of the proposed multimodal SAM-based tracking pipeline.
    \textbf{Step 1} shows the pre-training process of visual-geometric and visual-textual alignments. 
    \textbf{Step 2} shows the proposed multimodal teacher tracker pre-trained on visual, geometric, and language cues, while Step 3 shows the proposed unimodal student tracker distilling knowledge from the multimodal teacher tracker.}
    \label{fig_tracker}
    \end{minipage} 
    \vspace{-2em}
\end{figure*}

\textbf{Problem Formulation:}
The proposed MUTrack is based on a three-stage training paradigm as shown in Fig. \ref{fig_tracker}.
Given $B_1$ as initial bounding-box prompt in the first frame and underwater multi-modal video sequence, $\textbf{X}_{t} = \{\textbf{I}_t, \textbf{E}_t, \textbf{D}_t, \textbf{L}_t\}$,  where $\textbf{I}_{t}, \ = \{\textbf{I}_{t,i}\}_{i=1}^{n} \in \mathbb{R}^{h \times w \times 3}$ is underwater RGB sequence, $h$ and $w$ denote the spatial resolution and $n$ is the number of frames in that sequence.
$\textbf{E}_t$ is the corresponding enhanced RGB sequence, $\textbf{D}_t$ is the corresponding depth maps, $\textbf{L}_t$ is the corresponding textual description.
The main objective of MUTrack is to estimate the spatio-temporal state of a target in terms of its segmentation mask or bounding box across all frames. 

MUTrack is formulated based on the SAM2 framework \cite{kirillov2023segment} as a prompt-conditioned video segmentation approach, where temporal coherence is maintained through the memory encoder, memory attention, mask decoder, and memory bank modules. 
In MUTrack, we extend this formulation to a multi-modal learning during teacher training followed by KD in student network. 
The teacher network $T$ leverages multi-modal inputs to learn robust representations in marine environments. 
During inference, only the $I_{t}$ sequence and the initial bounding box $B_{1}$ are available.
The trained student network $S$ therefore performs $\textbf{M}_{t}= S(\textbf{I}_{1:n}, B_{1})$, tracking the target over time using features distilled from the multi-modal teacher.

\noindent \textbf{Stage I: Multimodal Alignment and Fusion} 

\noindent \textbf{Visual-Geometric Alignment:} Underwater imagery suffers from color attenuation and non-uniform illumination, making RGB-only features sensitive to photometric drift.
The objective of this stage is to align the enhanced RGB and depth encoders such that both modalities produce compatible feature representations.
To do so, we input $\textbf{E}_{t,i}$ and the corresponding $\textbf{D}_{t,i}$ into $E_r$ and $E_d$ encoders and get two feature representations $e_{t,i}$ and $d_{t,i} \in \mathbb{R}^{p_{r}}$.
We align both modalities using a symmetric contrastive loss as:

\begin{equation}
\mathcal{L}_{\text{cont}}^{d\to e}= - \frac{1}{K}\sum_{i=1}^{K}\log \Bigg ( \frac{\exp\big(\langle e_{t,i}, d_{t,i} \rangle / \tau \big)}
{\sum_{j=1}^{K} \exp\big(\langle e_{t,i}, d_{t,j} \rangle / \tau \big)} \Bigg), 
\end{equation}

\begin{equation}
\mathcal{L}_{\text{cont}}^{e\to d}= - \frac{1}{K}\sum_{i=1}^{K}\log \Bigg ( \frac{\exp\big(\langle d_{t,i}, e_{t,i} \rangle / \tau \big)}
{\sum_{j=1}^{K} \exp\big(\langle d_{t,i}, e_{t,j} \rangle / \tau \big)} \Bigg), 
\end{equation}

\noindent where $\tau$ is a temperature and $K$ is a batch size. The total loss is $\mathcal{L}_{cont}= \frac{1}{2}\big(\mathcal{L}_{cont}^{d\to e} + \mathcal{L}_{cont}^{e\to d}\big)$.

We also guide the depth encoder to match the enhanced RGB feature distribution via an $\ell_1$ regression with a stop-gradient on $e_{t,i}$:
\begin{equation}
\mathcal{L}_{\ell_{1}} \;=\; \frac{1}{|\Omega|}\sum_{i \in \Omega} \big\|\, d_{t,i} - \operatorname{sg}\!\big(e_{t,i}\big) \big\|_{1},
\end{equation}
where $\Omega$ indexes spatial locations.
The total pre-training loss is: $\mathcal{L}_{\text{pre}} = \mathcal{L}_{\ell_{1}} + \mathcal{L}_{cont}.$
After visual-geometric alignment, we fused the aligned visual and geometric features using element-wise max pooling $f^{vg}_{t,i}=\max(e_{t,i},d_{t,i})$ to be used as an input into the $T$ model.

\noindent \textbf{Visual-Textual Alignment:}
Using the language encoder $P_{l}$, we extract textual tokens $T_{t}$ from the description $L_{t}$ as:  $T_{t,1:n_t} = P(L_{t})$.
Using the prompt encoder $P_{e}$, we extract visual tokens $V_{t}$ using the bounding box $B_{1}$ as: $V_{t,1:n_v} = P_e(B_{1}, E_{t,1})$. 
We identify the top-$k$ positive visual-textual pairs ($T_{t,i}$, $V_{t,j}$), such that the similarity of these pairs is maximum over all pairs.
Using the vision to language adapter ($\Psi_{v2l}$), we learn a mapping of each visual token in the positive set to the corresponding language token as:
$\hat{T}_{t,i}=\Psi_{v2l} (V_{t,j})$, such that $\ell_{1}(\hat{T}_{t,i},{T}_{t,i})$ is minimized.
Finally, the aligned visual-textual tokens are concatenated to get the VL representation $f^{vl}_{t}$.

\noindent \textbf{Stage II: Multimodal Teacher:} The teacher network $T$ extends the SAM2 video object segmentation pipeline to a multi-modal settings. 
Given a visual-geometric fused representation $f^{vg}_{t,i}$ and VL  embedding $f^{vl}_{t}$ as input prompt, the teacher model produces frame-level mask predictions.
Particularly, the SAM2, consisting of memory attention, mask decoder, and memory encoder, is finetuned on MUOT-3M training set.
Due to multi-modal input samples, our teacher model encodes depth maps, language descriptions, and enhanced RGB frames information for improved UOT.

\noindent \textbf{Stage III: Unimodal Student Model:}  The unimodal student model $S$ is input with only underwater RGB frames and follows the SAM2 pipeline.
It mimics the multi-modal teacher model using four different KD losses. 
\textbf{i. Visual-Geometric KD:} This loss enforces alignment between the $S$ raw RGB features $f^{s}_{t,i}$ and the visual-geometric fused representation $f^{vg}_{t,i}$ as: $\mathcal{L}_{VG} = \frac{1}{hw}\left\|\,f^{s}_{t,i} - \operatorname{sg}\!\big(f^{vg}_{t,i}\big)\,\right\|_2^2$.
This loss is used to finetune the RGB encoder in the $S$ model to produce features similar to $f^{vg}_{t,i}$ using raw RGB frames. \textbf{ii. Spatiotemporal Attention Distillation:} The spatiotemporal teacher $T$ attention matrix $A^{T}_{t}$ in the SAM2 memory attention module captures long-range spatiotemporal dependencies among the fused visual-geometric features.
To enforce such dependencies in $S$, KD is performed between $T$ and $S$ attention maps $A^{S}_{t}$ as: $\mathcal{L}_{\text{A}} = 
\sum_{i=1}^{L} \left\| A_{t,i}^{T} - A_{t,i}^{S} \right\|_2^2$, where $L$ is the number of transformer layers in the memory attention network. \textbf{iii. VL Adapter Distillation:} The VL adapter in the teacher learns to map the visual prompts to the language prompts using enhanced RGB frames. 
In $S$, the same VL adapter is finetuned to map the visual prompts from the raw RGB frames to the prompts learned from the enhanced RGB frames as follows: $\mathcal{L}_{\text{VLA}} = \frac{1}{n_a}\left\|\,\hat{T}_{t,i} - \operatorname{sg}(\hat{T}^{S}_{t,i})\,\right\|_2^2$.
The learned prompts $\hat{T}^{S}_{t,i}$ capture the information contained in the language description while using raw RGB frames as input. \textbf{iv. Mask-Logit Distillation:} To transfer segmentation quality from the multimodal $T$ to unimodal $S$, we minimize the loss between the teacher $M_{t,i}^{T}$ and student masks $M_{t,i}^{S}$ logits: $\mathcal{L}_{\text{M}} =
\frac{1}{H'W'}\left\|\,M_{t,i}^{T}- \operatorname{sg}(M_{t,i}^{S})\,\right\|_2^2$.
As a result, the mask decoder in the $S$ is enabled to generate similar masks as the $T$ module while using raw RGB frames.
\noindent \textbf{Overall Loss:} The total KD objective is: $\mathcal{L}_{\text{KD}} = \mathcal{L}_{\text{VG}} + \mathcal{L}_{\text{A}} + \mathcal{L}_{\text{VLA}} + \mathcal{L}_{\text{M}}$. 
As a result, the student model in the MUTrack achieves robustness similar to the multi-modal teacher while using only raw RGB frames as input.
\vspace{-1em}



\noindent 



%% file: sec/results.tex
\section{Experiments}
\label{sec:results}
\textbf{5.1.~Training and Implementation Details:} All experiments are conducted using PyTorch on a workstation equipped with 2 RTX A6000 GPUs (48 GB each) and Intel Xeon Platinum 8358 CPUs.
We finetuned the visual encoder of CLIP (ViT-B/224) as our RGB, depth, and prompt encoders, and textual encoders as the language encoder.  
All input frames are resized and normalized to $1024 \times 1024$ pixels \cite{deng2009imagenet}. 
Each training batch include consecutive frames randomly sampled from video clips to preserve temporal consistency.
In vision-geometric alignment, we finetuned depth and RGB encoders for 50 epochs with batch size 16 and learning rate $10^{-4}$ (AdamW, $\beta_{1}$ = 0.9, $\beta_{2}$ = 0.999, weight decay = $10^{-2}$).
In VL alignment, the same settings are used for training adapter consisting of two layers fully connected network and finetuning of prompt encoder.
A multi-modal $T$ is finetuned for 60 epochs with batch size 8 and base learning rate $5 \times 10^{-5}$ (cosine decay) employing SAM2 loss \cite{kirillov2023segment}.
A unimodal $S$ is trained independently for 30 epochs with batch size 8 and base learning rate $5 \times 10^{-5}$ (cosine decay) using raw RGB frames only.
We pre-trained our four KD losses in this stage to mimic the multimodal $T$.
During inference, raw RGB-only $S$ is used.
We split the proposed MUOT-3M dataset into 70$\%$ training videos (containing 2.10 million frames) and 30$\%$ testing videos (0.90 million frames).
\textit{MUTrack operates in real time at 24 FPS.} \\
\noindent \textbf{SOTA Trackers:} We evaluated 23 SOTA trackers on our MUOT-3M dataset.
SOTA are VL-based trackers: ATCTrack \cite{ATCTrack}, DUTrack \cite{DUTrack}, CiteTracker\cite{citetracker}, and JointNLT\cite{JointNL}.
DCFs-based trackers: ATOM \cite{ATOM}, and ToMP \cite{ToMP}.
Siamese-based trackers: SIMTrack \cite{SIMTrack}, AQATrack \cite{AQATrack}, ARTrack \cite{ARTrack}, and ARTrackv2 \cite{ARTrackV2}.
ViT-based trackers: CompressTrack \cite{CompressTrack}, CSWinTTrack \cite{CSWinTT}, DropTrack \cite{DropTrack}, HiT \cite{HiT}, HIPTrack \cite{HIPTrack}, OSTrack \cite{OSTrack}, SeqTrack \cite{SeqTrack}, GRM \cite{GRM}, STARK \cite{STARK}, SuperSBT \cite{SuperSBT}, UOSTrack \cite{li2023underwater}, TransT \cite{TransT}, and ZoomTrack \cite{ZoomTracck}. \\
\noindent \textbf{5.2.~Evaluation Protocols} \\
\textbf{Protocol I: Cross-Domain Evaluation:} In this protocol, we perform a cross-domain evaluation to assess the generalization capability of SOTA trackers trained on terrestrial datasets and tested on MUOT-3M test set. \noindent \textbf{Protocol II: Underwater Domain Evaluation:} It focuses on underwater domain-specific evaluation to establish a fair and unified benchmark for the UOT community.
A set of representative terrestrial trackers, including ATCTrack \cite{ATCTrack}, DUTrack \cite{DUTrack}, ARTrack \cite{ARTrack}, ARTrackV2 \cite{ARTrackV2}, and SuperSBT \cite{SuperSBT}, are finetuned and tested using the training/testing splits of MUOT-3M dataset. 
\begin{table}[t!]
\resizebox{\linewidth}{!}{%
\centering
\begin{tabular}{l|c c cc|c cc|}
\hline
Multimodal &  \multicolumn{4}{c|}{Modalities} &MUOT-3M &WebUOT-1M \\
Teacher (MMT)&\textbf{I}&\textbf{E}&\textbf{D}&\textbf{L}&(success Rate)&(success Rate)\\
\hline
Baseline SAM2 \cite{kirillov2023segment} &$\checkmark$ &$\times$&$\times$ &$\times$&60.12&61.11\\
Baseline SAM2 \cite{kirillov2023segment} &$\times$ &$\checkmark$&$\times$ &$\times$&61.20&62.23\\
Baseline SAM2 \cite{kirillov2023segment} &$\times$ &$\times$&$\checkmark$ &$\times$&58.55&59.21\\
\hline
MUTrack-MMT$_{2}$&$\times$ &$\checkmark$&$\checkmark$ &$\times$&66.12&65.59\\
MUTrack-MMT$_{3}$&$\times$ &$\checkmark$&$\times$ &$\checkmark$&65.70&63.66\\
MUTrack-MMT$_{4}$&$\times$ &$\times$&$\checkmark$ &$\checkmark$&66.50&64.56\\
MUTrack-MMT$_{5}$& $\checkmark$ &  $\times$&$\checkmark$ &$\checkmark$&\underline{67.70}&\underline{66.93}\\
\hline
MUTrack-MMT&$\times$ &  $\checkmark$&$\checkmark$ &$\checkmark$&\textbf{68.10}&\textbf{68.79}\\
\hline
\end{tabular}
}
\vspace{-1em}
\caption{MUTrack teacher-only results with baseline SAM tracker. MMT stands for ``\textit{Multimodal Teacher}''.}
\label{table_modalities}
\vspace{-1em}
\end{table}
\begin{table}[t]
\resizebox{\linewidth}{!}{%
\centering
\begin{tabular}{l|c|c c cc|c cc|}
\hline
Unimodal& Input& \multicolumn{4}{c|}{KD Losses} &MUOT-3M &WebUOT-1M\\
Student (UMS)&&$\mathcal{L}_{VG}$&$\mathcal{L}_{A}$&$\mathcal{L}_{VLA}$&$\mathcal{L}_{M}$&(Success Rate)&(Success Rate)\\
\hline
MUTrack-UMS &\textbf{I}&$\checkmark$ &  $\checkmark$&$\checkmark$ &$\checkmark$&\underline{66.58}&\underline{67.10}\\
\hline
MUTrack-UMS$_{1}$&\textbf{I}&$\checkmark$ &$\checkmark$&$\checkmark$ &$\times$&64.22&65.31\\
MUTrack-UMS$_{2}$&\textbf{I}&$\checkmark$ &$\checkmark$&$\times$ &$\checkmark$&64.10&63.96\\
MUTrack-UMS$_{3}$&\textbf{I}&$\checkmark$ &$\times$&$\checkmark$ &$\checkmark$&64.50&66.09\\
MUTrack-UMS$_{4}$&\textbf{I}&$\times$ &$\checkmark$&$\checkmark$ &$\checkmark$&63.16&64.77\\
MUTrack-UMS$_{5}$&\textbf{E}&$\checkmark$ &  $\checkmark$&$\checkmark$ &$\checkmark$&\textbf{67.20}&\textbf{68.30}\\
\hline
\end{tabular}
}
\vspace{-1em}
\caption{Teacher is fixed and MUTrack student-only results are reported. UMS stands for “\textit{Unimodal Student}”.}
\label{table_KD}
\vspace{-1em}
\end{table}

\begin{table*}[t]
\begin{minipage}{0.70\textwidth}
   \centering \small
\setlength{\tabcolsep}{2.0pt}
\centering
   \scalebox{0.80}{
\begin{tabular}{l|ccc|ccc|ccc|ccc|ccc|}
\hline
\multirow{2}{*}{Trackers}&\multicolumn{3}{c|}{MUOT-3M} &\multicolumn{3}{c|}{WebUOT-1M}&\multicolumn{3}{c|}{UTB180}&\multicolumn{3}{c|}{UVOT400}&\multicolumn{3}{c|}{UWCOT22}\\\
&S&P&NP&S&P&NP&S&P&NP&S&P&NP&S&P&NP\\
\hline
MUTrack (MMT)&\textbf{68.10}&\textbf{69.93}&\textbf{86.66}&\textbf{68.79}&\textbf{70.33}&\textbf{88.20}&\textbf{73.77}&\textbf{74.33}&\textbf{81.69}&\textbf{72.10}&\textbf{75.87}&\textbf{80.67}&\textbf{79.40}&\textbf{81.66}&\textbf{83.22}\\
MUTrack (UMS)&\underline{66.58}&\underline{68.16}&\underline{84.32}&\underline{67.10}&\underline{69.33}&\underline{86.97}&\underline{72.88}&\underline{76.61}&\underline{80.30}&\underline{70.55}&\underline{74.51}&\underline{78.97}&\underline{77.87}&\underline{80.10}&\underline{81.11}\\
UOSTack&49.22&41.92&61.20   &53.14&47.76&65.67   &52.67&45.87&68.22 &44.25&39.67&56.32& 45.31&35.95&53.30\\
DUTrack&62.66&60.20&79.90        &63.81&64.27&81.70   &64.61&62.50&83.73 &64.16&63.96&80.98& 62.08&59.23&75.85\\
ATC&62.32&60.37&76.50       &64.66&65.48&79.10   &64.16&65.11&79.43 &61.58&63.08&74.47& 62.58&61.03&73.93\\
ARv2&58.22&53.69&71.81      &61.18&59.73&75.40   &61.34&58.48&76.52 &58.60&57.56&71.16& 57.97&52.24&67.18\\
SuperSBT&55.77&41.92&61.20  &58.35&56.94&72.33   &58.09&56.38&74.12 &55.49&55.74&68.59& 54.84&49.28&64.88\\
\hline
\end{tabular}
}
\end{minipage}
  \hfill
  \begin{minipage}{0.28\textwidth}
  \small 
\caption{\textbf{Cross-dataset performance comparison of SOTA trackers:} 
MMT stands for ``\textit{Multimodal Teacher}'', UMS is ``Unimodal Student'', S is  ``Success rate'', P is ``Precision'' and NP is ``Normalized Precision''.
} 
\label{table_generaliation}
 \end{minipage}
\end{table*}

\begin{table*}[t]
\begin{minipage}{0.70\textwidth}
   \centering \small
\setlength{\tabcolsep}{2.0pt}
\centering
   \scalebox{0.80}{
\begin{tabular}{l|ccc|ccc|ccc|ccc|ccc|}
\hline
\multirow{2}{*}{Trackers}&\multicolumn{3}{c|}{MUOT-3M} &\multicolumn{3}{c|}{WebUOT-1M}&\multicolumn{3}{c|}{UTB180}&\multicolumn{3}{c|}{UVOT400}&\multicolumn{3}{c|}{UWCOT22}\\\
&S&P&NP&S&P&NP&S&P&NP&S&P&NP&S&P&NP\\
\hline
MUTrack (MMT)&68.10&69.93&86.66&\underline{68.79}&70.33&\underline{88.20}&\underline{73.77}&74.33&81.69&72.10&75.87&80.67&79.40&81.66&83.22\\
MUTrack (UMS)&66.58&68.16&84.32&67.10&69.33&86.97&72.88&\textbf{76.61}&80.30&70.55&74.51&78.97&77.87&80.10&81.11\\
\hline
DAM4SAM \cite{videnovic2025distractor}&62.33&64.21&75.12&61.10&62.34&70.23&68.10&70.84&76.66&68.97&69.81&74.44&71.12&73.20&80.12\\
MUTrack (MMT-DAM4SAM)&70.13&71.23&87.33&\textbf{69.01}&\textbf{72.43}&\textbf{88.32}&\textbf{74.56}&75.14&\textbf{83.61}&\textbf{74.56}&\underline{76.44}&\underline{82.34}&\underline{80.12}&\textbf{82.08}&\underline{84.61}\\
MUTrack (UMS-DAM4SAM)&68.69&\underline{70.32}&85.13&66.10&\underline{71.40}&86.56&73.65&74.46&82.69&73.65&75.14&81.63&79.21&81.80&82.29\\
\hline
SAMURAI \cite{yang2024samurai}&61.31&62.50&73.21&58.88&61.44&69.32&66.01&69.48&75.16&65.79&67.18&72.44&70.70&71.10&78.33\\
MUTrack (MMT-SAMURAI)&67.31&69.32&\textbf{88.31}&67.10&70.13&85.23&73.66&74.41&\underline{82.60}&\underline{73.50}&\textbf{77.48}&\textbf{83.30}&\textbf{81.21}&\underline{81.80}&\textbf{85.15}\\
MUTrack (UMS-SAMURAI)&67.10&69.12&\underline{88.10}&66.09&68.31&84.32&72.61&74.10&81.06&72.05&75.88&81.03&\underline{80.12}&80.10&83.51\\
\hline
SAM2Long \cite{ding2025sam2long}&61.11&63.11&76.39&62.01&61.44&69.32&67.01&71.48&75.16&67.79&66.10&71.43&70.21&71.02&79.21\\
MUTrack (MMT-SAM2Long)&\textbf{71.31}&\textbf{72.32}&86.13&67.10&71.34&86.23&73.65&\underline{76.41}&82.16&72.65&75.41&81.43&79.21&80.80&83.16\\
MUTrack (UMS-SAM2Long)&\underline{70.96}&70.12&84.31&65.01&69.13&85.65&71.51&75.64&81.96&71.58&74.44&80.33&78.12&79.89&81.92\\
\hline
\end{tabular}
}
\end{minipage}
  \hfill
  \begin{minipage}{0.20\textwidth}
  \small 
\caption{\textbf{MUTrack generalization with SOTA Trackers:} Our MMT and UMS are plugged into the recent SOTA SAM-based tracking pipelines.
MMT stands for ``\textit{Multimodal Teacher}'', UMS is ``Unimodal Student'', S is  ``Success rate'', P is ``Precision'' and NP is ``Normalized Precision''.
} 
\label{table_sota_trackers}
\vspace{-1em}
 \end{minipage}
\end{table*}
\begin{figure}[t!]
    \centering
    \begin{subfigure}[t]{0.15\textwidth}
        \centering
        \includegraphics[width=\textwidth]{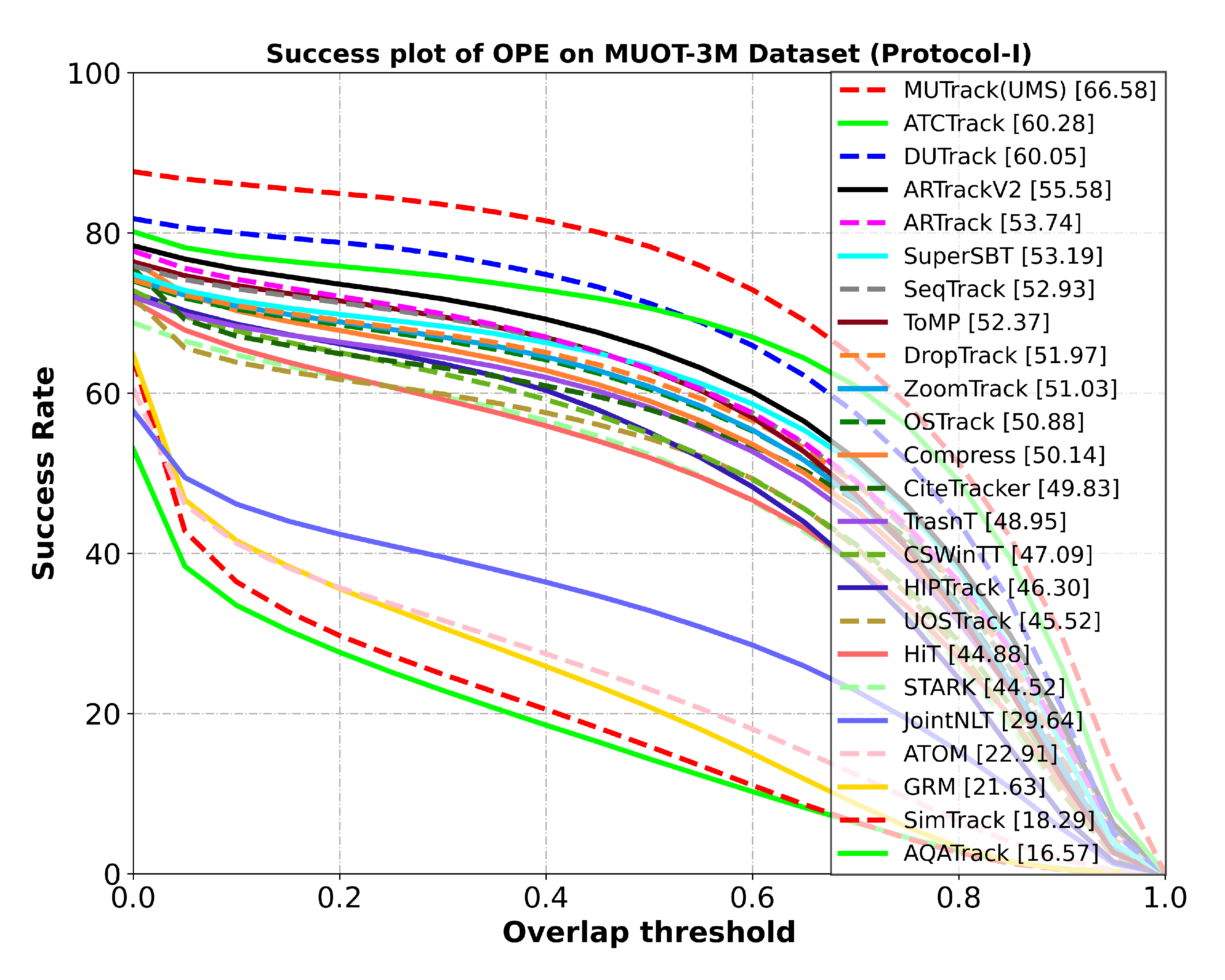}
        \caption{Success Plot}
        \label{fig:plot1}
    \end{subfigure}
    \hfill
    \begin{subfigure}[t]{0.15\textwidth}
        \centering
        \includegraphics[width=\textwidth]{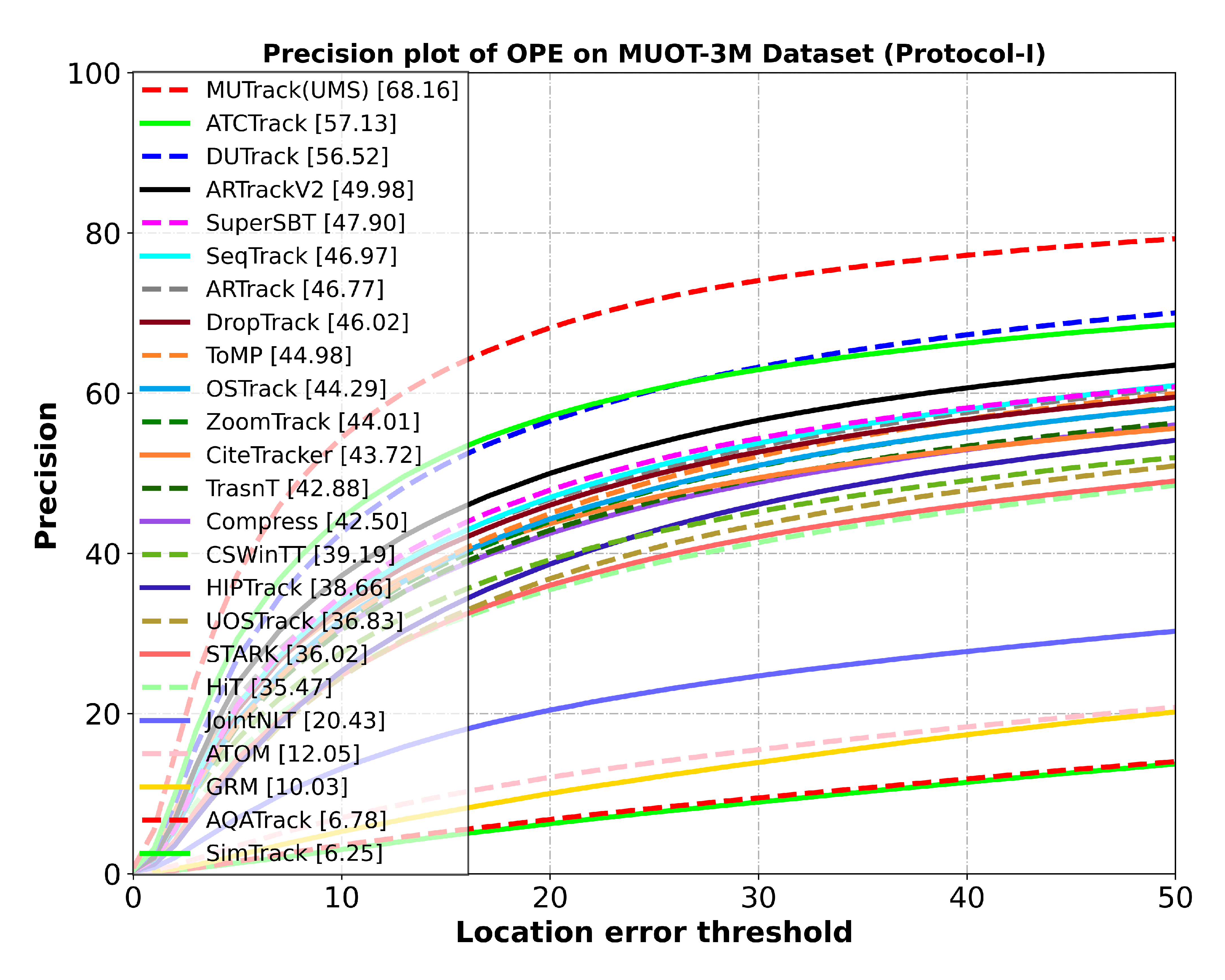}
        \caption{Precision Plot}
        \label{fig:plot2}
    \end{subfigure}
    \hfill
    \begin{subfigure}[t]{0.15\textwidth}
        \centering
        \includegraphics[width=\textwidth]{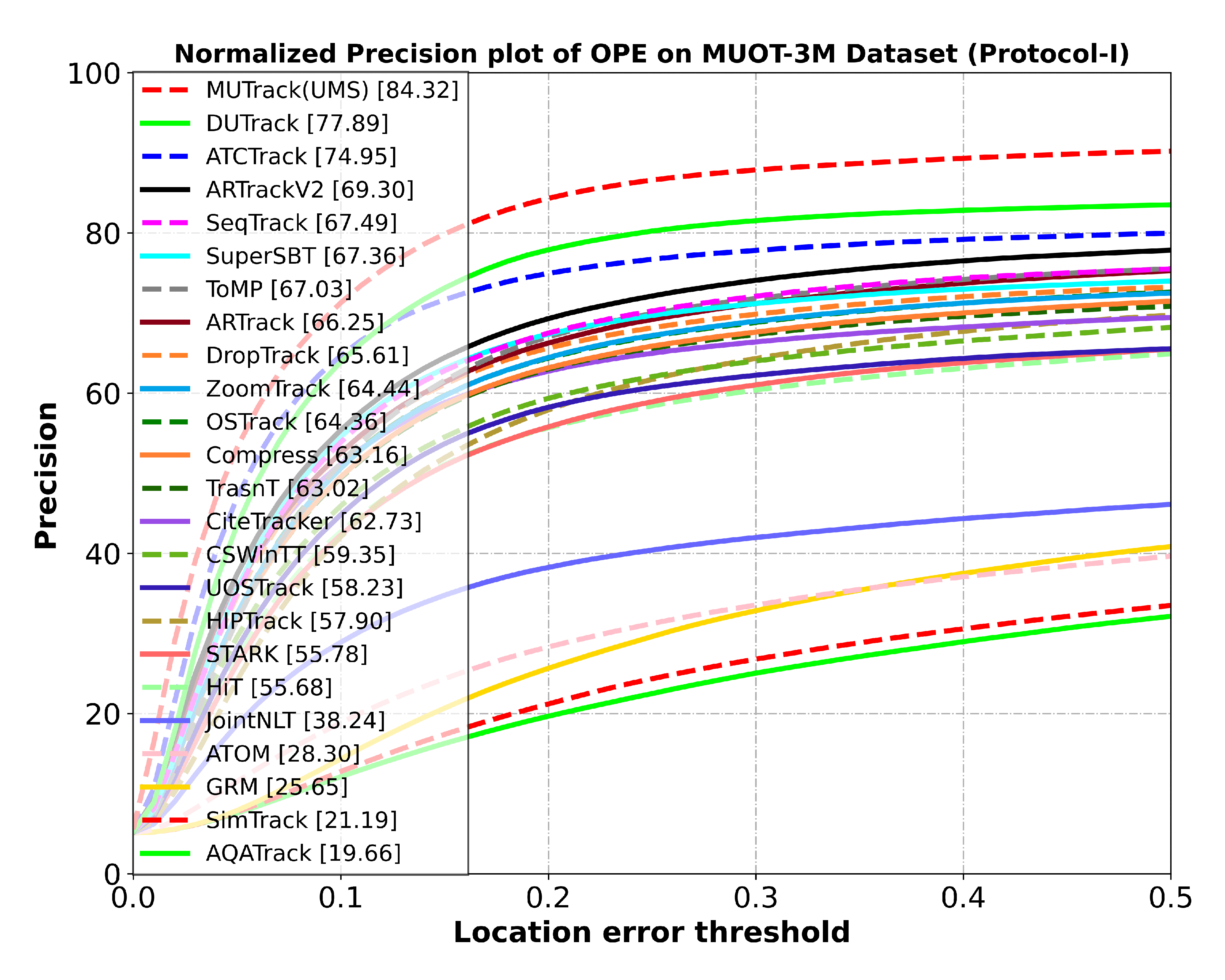}
        \caption{Norm. Precision}
        \label{fig:plot3}
    \end{subfigure}
    \begin{subfigure}[t]{0.15\textwidth}
        \centering
        \includegraphics[width=\textwidth]{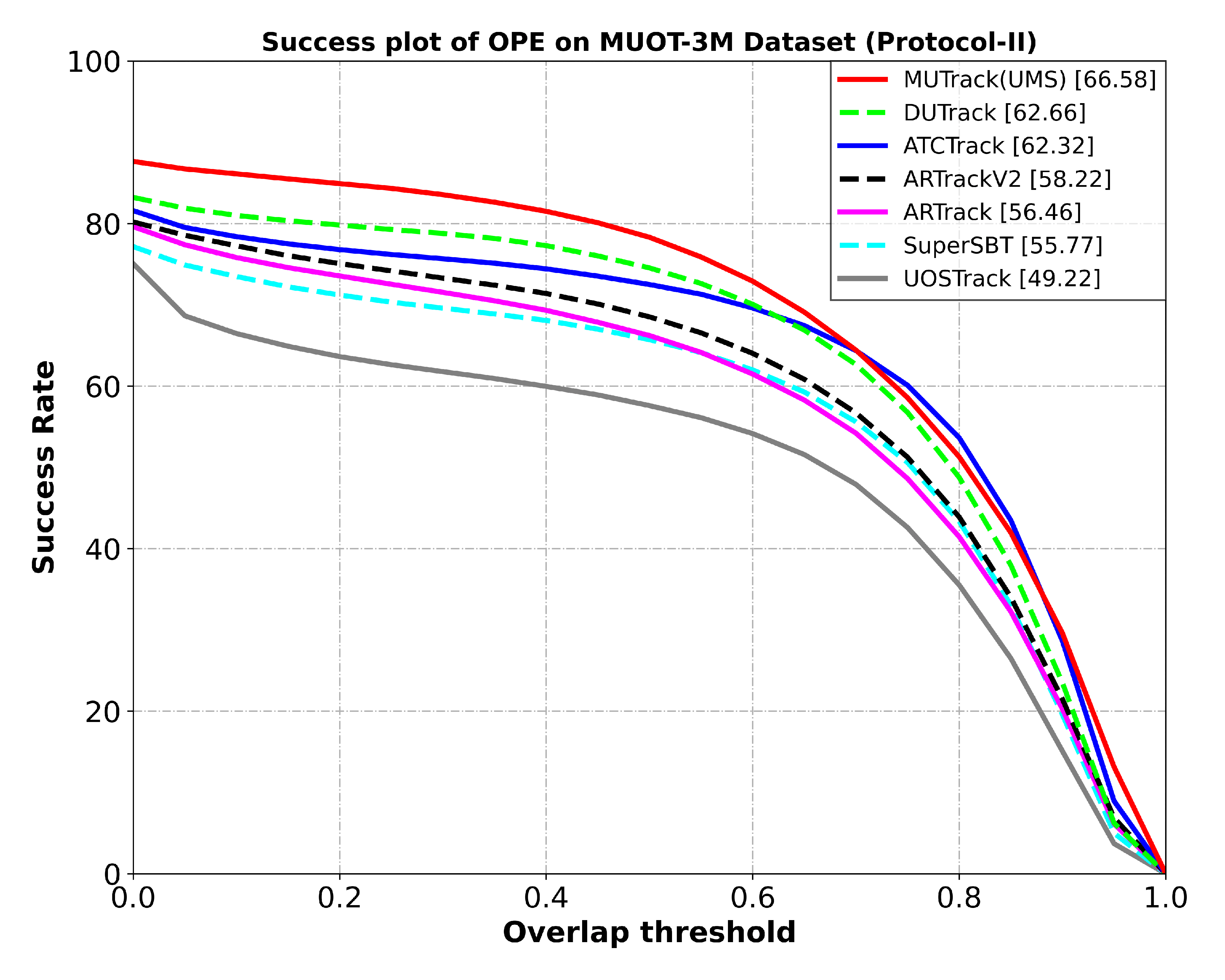}
        \caption{Success Plot}
        \label{fig:plot4}
    \end{subfigure}
    \hfill
    \begin{subfigure}[t]{0.15\textwidth}
        \centering
        \includegraphics[width=\textwidth]{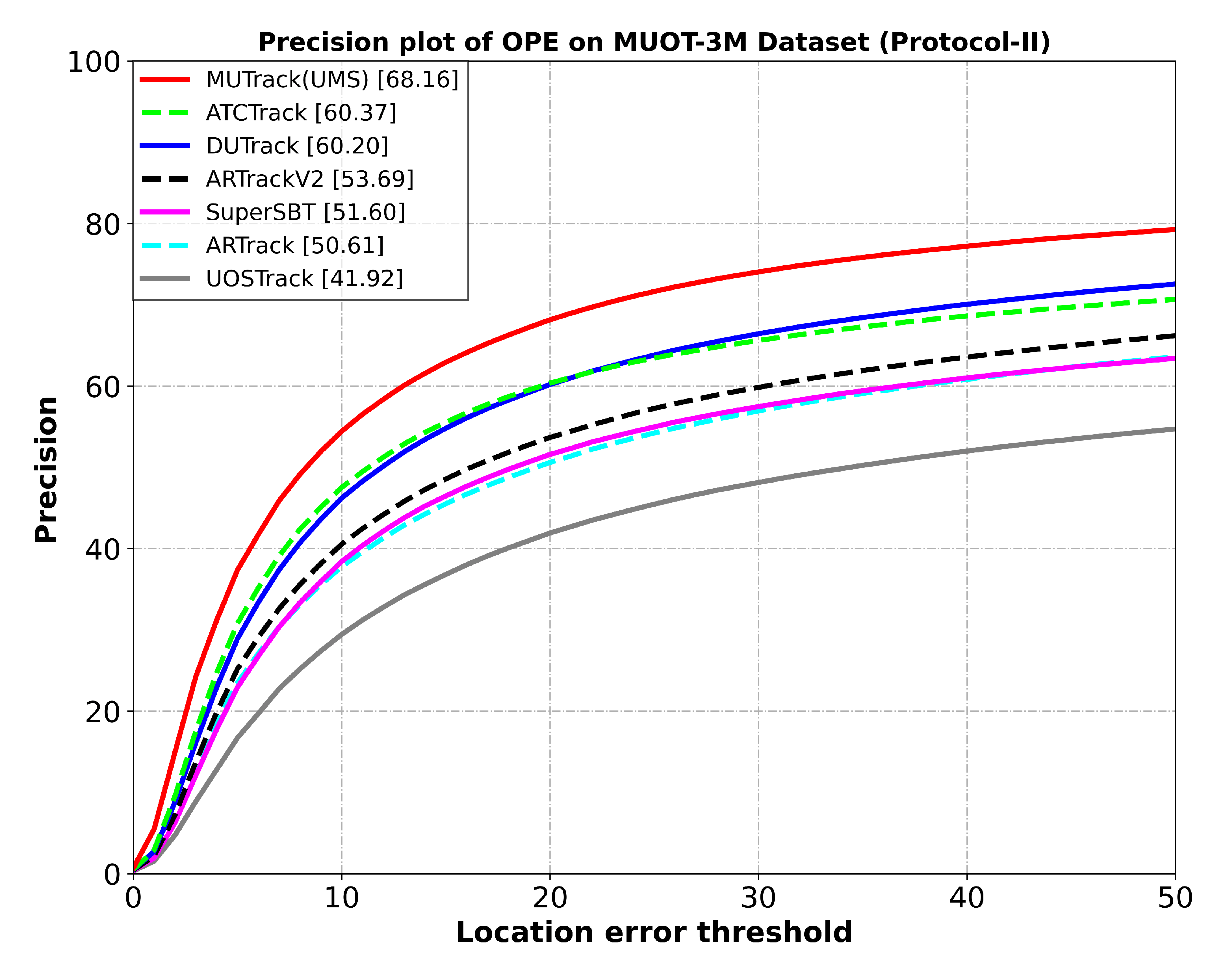}
        \caption{Precision Plot}
        \label{fig:plot5}
    \end{subfigure}
    \hfill
    \begin{subfigure}[t]{0.15\textwidth}
        \centering
        \includegraphics[width=\textwidth]{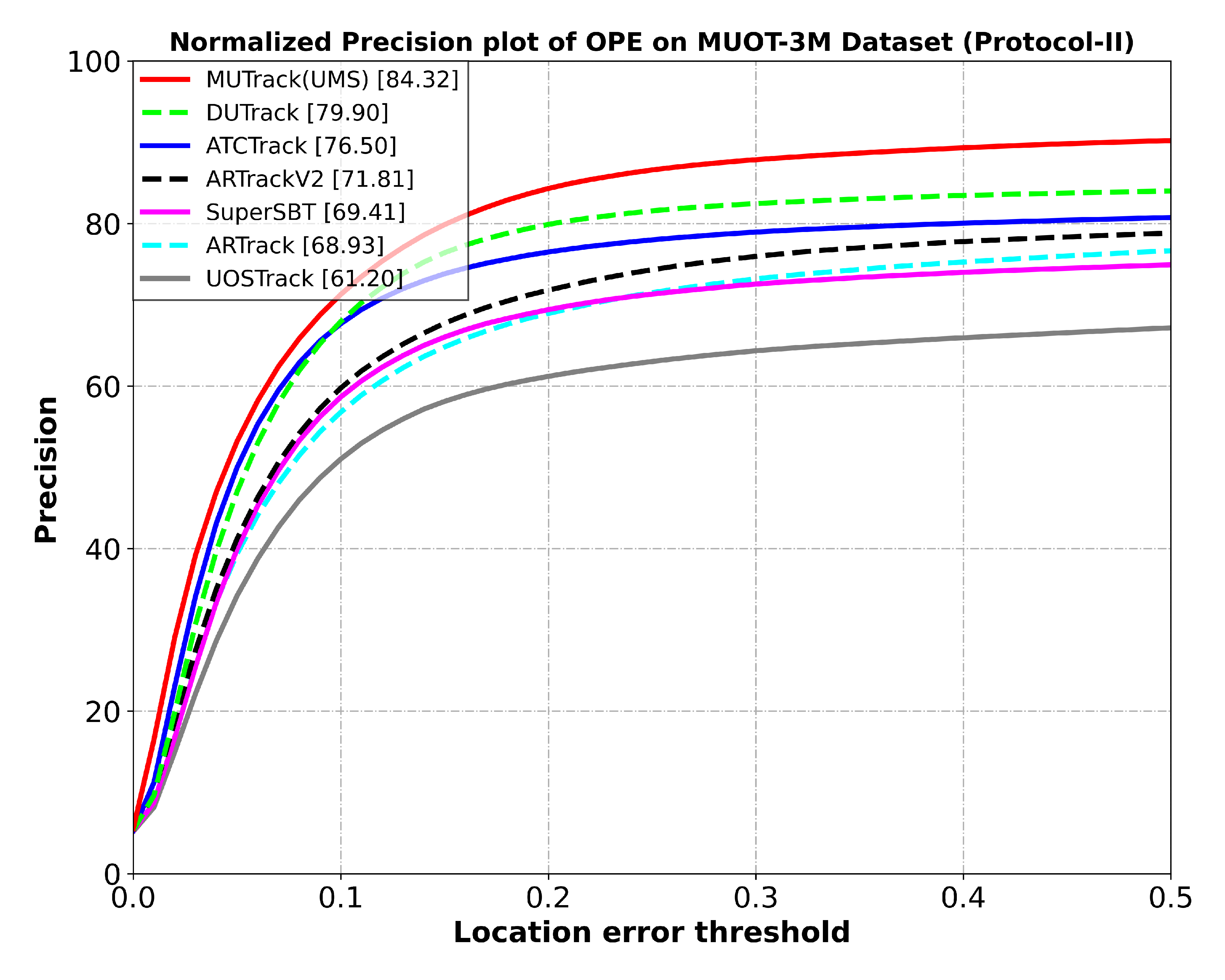}
        \caption{Norm. Precision}
        \label{fig:plot6}
    \end{subfigure}
    \vspace{-1em}
    \caption{Performance comparison of MUTrack with SOTA trackers using protocol \textbf{I} and \textbf{II} on MUOT-3M dataset. Norm. stands for ``Normalized''.}
    \label{fig:sixplots}
    \vspace{-2em}
\end{figure}
\noindent \textbf{5.3.~Comparison with SOTA Trackers:} The SOTA trackers are compared with our proposed MUTrack (unimodal student tracker) using the testing split of MUOT-3M in terms of success, precision, and normalized precision (Fig. \ref{fig:sixplots}). 
The performance of SOTA trackers using protocol \textbf{I} has remained lower as compared to the performance in protocol \textbf{II} (Figs. \ref{fig:sixplots} (a)-(f)). 
It is because of the retraining of five SOTA trackers on the MUOT-3M training split.
The proposed MUTrack has obtained a success rate of 66.58$\%$, which is \textbf{3.92}$\%$ better than the second performing tracker DUTrack, obtaining \textbf{62.66}$\%$.   
In terms of precision, MUTrack has obtained \textbf{68.16}$\%$, which is \textbf{7.79}$\%$ better than the second best performer, ATC.
In terms of normalized precision, MUTrack has obtained \textbf{84.32}$\%$ which is \textbf{4.42}$\%$ better than DUTrack. 
The performance improvement in MUTrack underscores the significant importance of multimodal alignment and fusion in the $T$ network and multi-level KD used to train the unimodal $S$ tracker.\\
\noindent \textbf{5.4.~Cross-Dataset Results:} We compare the performance of MUTrack (both MMT and UMS) with the six SOTA trackers using a training split of MUOT-3M and testing split of five existing underwater datasets using Protocol \textbf{II} (Table \ref{table_generaliation}). 
We observed that the proposed MUTrack has consistently outperformed all SOTA trackers and demonstrated better generalization capabilities across all datasets.
\textit{This experiment underscores the significance of the proposed MUTrack architecture, consisting of multimodal teacher training and unimodal student testing}.\\
\noindent \textbf{5.5.~MUTrack Generalization with SOTA Trackers:} To evaluate the generalization capability of MUTrack, we integrated its multimodal $T$ and unimodal $S$ modules into three recently proposed SOTA SAM-based trackers, including DAM4SAM \cite{videnovic2025distractor}, SAM2Long \cite{ding2025sam2long}, and SAMURAI \cite{yang2024samurai}.
Table~\ref{table_sota_trackers} shows the results using the test splits of MUOT-3M and WebUOT-1M datasets.
\textit{Across all three baselines, integrating MuTrack consistently improves tracking performance.}\\
\noindent \textbf{5.6.~Attribute-based Performance:} \textit{Please see supplementary material.}
\\
\textbf{5.7.~Ablation Studies} \\
\textbf{1. Multimodal Teacher (MMT)  Vs. Unimodal Student (UMS) Tracker (Tables \ref{table_modalities}-\ref{table_KD}):} The proposed MMT tracker has consistently shown better performance than the baseline SAM2 and proposed UMS trackers, as shown in Tables \ref{table_modalities} \& \ref{table_KD}.
Specifically, on the MUOT-3M test split, the performance degradation is xyz$\%$ and abc on WebUOT-1M.
This demonstrates the effectiveness of our four-level KD strategies that efficiently transfer knowledge from the MMT tracker to the UMS tracker.
Due to the simple architecture of our UMS tracker, it processes 28 fps compared to our MMT tracker that processes abc fps.
The reduced performance of our UMS tracker is justified due to the reduced data requirements and fast inference speed. \\
\noindent \textbf{2. Variation of Modalities and Corresponding Loss Functions (Tables \ref{table_modalities}):} In Table \ref{table_modalities}, we compare the performance of proposed Multimodal Teacher Tracker (MMT) using three modalities including enhanced RGB (\textbf{E}), depth map (\textbf{D}), and language (\textbf{L}) with reduced modalities.  
In MUTrack-MMT$_{1}$, we keep only \textbf{E} resulting in no multi-modal alignment and no fusion in the teacher tracker.
It only learns from the enhanced RGB images.
In MUTrack-MMT$_{2}$, we keep only \textbf{E} and \textbf{D}, resulting in the removal of VL alignment while visual-geometric is used.
In MUTrack-MMT$_{3}$, we keep only \textbf{E} and \textbf{L}, resulting in the removal of vision-geometric alignment while VL alignment is used.
In MUTrack-MMT$_{4}$, we keep only \textbf{D} and \textbf{L}, resulting in the removal of both multi-modal alignments where the input features are computed by the depth encoder and only textual prompts are used. \textit{We observed performance degradation in all cases when the modalities are reduced from the three proposed \textbf{E}, \textbf{D}, and \textbf{L}, demonstrating the importance of each modality in both UOT datasets.
In MUTrack-MMT$_{5}$, we replace \textbf{E} with raw RGB frames \textbf{I} along with \textbf{D} and \textbf{L}, resulting in performance degradation, underscoring the need for \textbf{E} during teacher tracker training.} \\
\noindent \textbf{3. Variations of Knowledge Distillation (KD)  Losses in Unimodal Student (UMS) Tracker (Table \ref{table_KD}):} In Table \ref{table_KD}, we evaluate the significance of each KD loss one by one used during the student tracker and observed the performance variation.
We observed a performance degradation in each case, demonstrating the importance of each KD loss.
In MUTrack-UMS$_{5}$, we replace input \textbf{I} with \textbf{E}, resulting in a performance improvement; however, with a cost of computing \textbf{E} modality during inference.

%% file: sec/Conclusion.tex
\section{Conclusion \& Future Directions}
\label{sec:conclusion}
\vspace{-1em}
We proposed a large-scale underwater visual object tracking dataset, MUOT-3M, consisting of 3M frames and a novel underwater visual tracker, MUTrack, utilizing  RGB, depth, and language modalities. 
MUOT-3M consists of 3030 underwater video sequences, which are obtained after the filtering process.
It is densely annotated with bounding boxes and a segmentation mask consisting of 32 tracking attributes, 16 Phylum categories, 124 families, and 677 fine-grained classes.
MUTrack consists of a SAM2-based Multimodal Teacher (MMT) and a Unimodal Student (UMS) tracker.
The MMT contains multimodal alignment and fusion followed by fine-tuning of the SAM2 model for multimodal input and prompts.
We proposed visual-geometric alignment and fusion using enhanced RGB and depth frames for computing input features for SAM, as well as visual-language alignment and fusion for computing multimodal prompts. 
For training UMS, we proposed four-level knowledge distillation losses that transfer the multi-modality information from the teacher to the student.
Our proposed UMS takes only raw RGB frames as input and learns a performance equivalent to enhanced RGB frames, depth maps, and language modality.
We evaluate and compare our MUTrack on two large-scale datasets, including MUOT-3M and WebUOT-1M.
We also evaluate the generalization ability of MUTrack-3M on the other four publicly available UOT datasets.
Results demonstrate the superior performance of MUTrack compared to the SOTA trackers.

%% file: supplementary.tex
\begin{figure*}[htp]
\centering
\includegraphics[width=\linewidth]{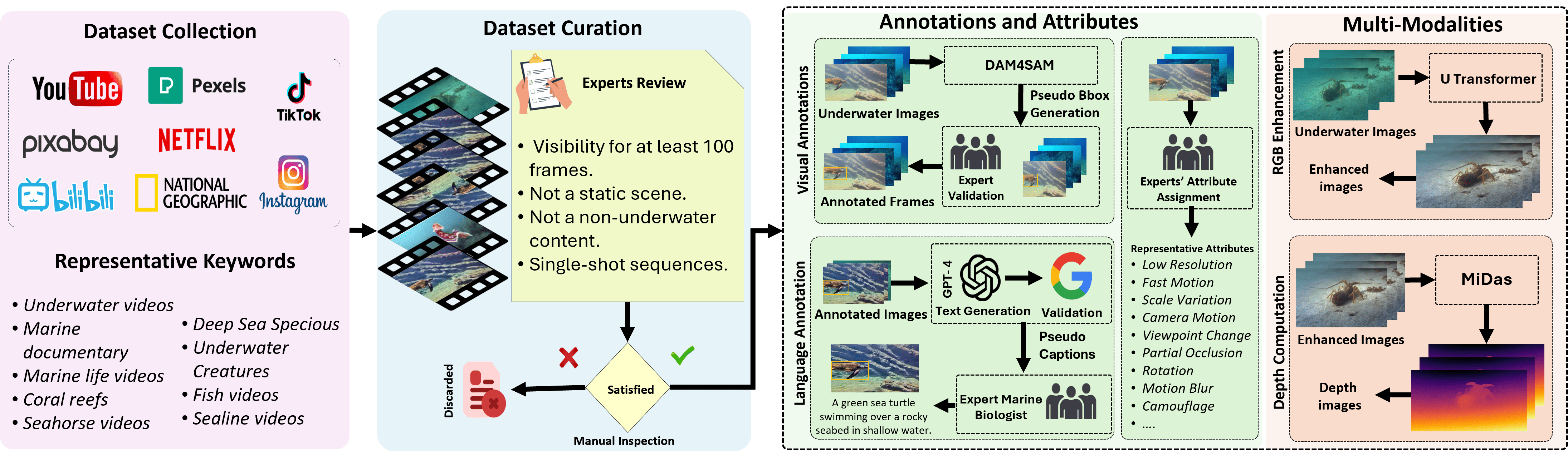}
\caption{Overview of the MUOT-3M dataset collection and curation pipeline. The process includes three main stages: large-scale video collection from multiple sources, expert-driven curation to ensure underwater authenticity and visual quality, and annotation with multi-modal data generation. Each stage is standardized to maintain consistency and ensure high-quality tracking sequences across diverse marine environments.}

\label{Collection}
\end{figure*}

\section{MUOT-3M Dataset Construction}
\label{dataset}

The MUOT-3M dataset was built through a structured pipeline integrating large-scale video collection, expert curation, and multi-modal data generation. As shown in Fig.~\ref{Collection}, the process includes three stages: \textit{collection}, \textit{curation}, and \textit{annotation with modality generation}. Each stage ensures visual diversity, annotation accuracy, and ecological balance across marine environments. The pipeline combines automated preprocessing with manual verification to maintain consistency and eliminate noise. All stages were standardized under a unified protocol to enable reproducible dataset construction and quality control.

\subsection{Dataset Collection}

We aggregated underwater footage from multiple open-source and social media platforms, including YouTube, Pexels, TikTok, Pixabay, Netflix, BiliBili, National Geographic, and Instagram. These platforms provide a rich mix of professional documentary content and user-generated videos that capture a wide variety of species, lighting conditions, camera perspectives, and motion patterns.

The collection process prioritized both scientific and practical diversity. Videos were retrieved using targeted search keywords covering marine biodiversity, habitat types, and environmental challenges. For each source, we filtered for high-definition clips containing continuous underwater motion and natural lighting. Videos with artificial overlays, excessive editing, or duplicated content were removed.

\subsection{YouTube Marine Videos}

YouTube served as a primary source of long underwater sequences, including professional documentaries and research footage. Videos recorded with stable camera platforms were prioritized to support precise bounding-box annotation. Representative keywords included \textit{``underwater videos,'' ``marine documentary,'' ``marine life videos,'' ``coral reefs,'' ``deep sea exploration,'' ``underwater creatures,''} and \textit{``oceanic wildlife.''} Only single-shot clips with continuous underwater visibility and motion lasting over 100 frames were retained.

\subsection{BiliBili Underwater Videos}

BiliBili provided additional underwater recordings from different regions, including diver and research footage. Keywords used included \textit{``ocean exploration,'' ``underwater world,'' ``deep sea species,'' ``coral reef fish,'' ``marine ecosystem,'' ``seahorse,''} and \textit{``diving videos.''} Clips showing surface scenes, static cameras, or non-marine content were excluded. The selected videos captured diverse marine species under varying water clarity and illumination.

\subsection{Pexels and PixaBay Videos}

Pexels and Pixabay provided high-quality open-source underwater videos with stable motion and clear visuals. Keywords included \textit{``underwater animals,'' ``marine biodiversity,'' ``sea turtle,'' ``reef footage,'' ``Dolphins,'' ``shark videos,''} and \textit{``ocean floor.''} These license-free clips offered controlled lighting and water conditions suitable for baseline evaluation and bounding-box generation.

\subsection{Netflix and National Geographic Videos}

Netflix and National Geographic provided long-form professional underwater footage covering coral reefs, open ocean, and deep-sea environments. From each video, we trimmed single-shot underwater segments with continuous motion and no transitions. Segments meeting visibility and underwater validity criteria, with a minimum of 100 frames, were retained. These sources offered stable, high-resolution content that improved temporal continuity and reduced editing artifacts, enhancing the dataset’s reliability for tracking evaluation.

\subsection{Social Media Platform Videos}

Short-form videos from TikTok and Instagram were collected to include in-the-wild underwater scenes. These clips often contained handheld motion, fast-moving targets, and variable lighting. Searches used hashtags and captions such as \textit{\#underwatervideos, \#scubadiving, \#sealife, \#coralreef, \#marinelife, \#deepsea,} and \textit{\#underwaterphotography.} This source increased the diversity and realism of motion patterns, improving dataset robustness in unconstrained tracking conditions.

\subsection{Dataset Curation}

All collected clips underwent manual review to ensure quality and authenticity. A team of marine biologists and computer vision researchers inspected each sequence for visibility, motion continuity, and underwater validity. The following criteria guided selection:

\begin{itemize}
    \item \textbf{Visibility:} The target must remain visible for at least 100 consecutive frames.
    \item \textbf{Scene dynamics:} The sequence must show continuous motion; static or looped clips were excluded.
    \item \textbf{Underwater authenticity:} Non-underwater content was removed. A limited number of synthetic or animation based clips were retained to enhance diversity and evaluate model robustness under artificial conditions.
    \item \textbf{Single-shot requirement:} Only uncut clips without transitions were retained.
\end{itemize}

This process ensured a balanced dataset containing natural and controlled scenarios while maintaining consistency and visual quality across all sequences. Video clips failing these conditions were replaced through iterative searching. The remaining videos were checked for frame consistency and duplication. This two-stage curation ensured a balanced dataset with diverse marine scenes and stable visual quality.

\begin{figure*}
\centering
\includegraphics[width=0.85\linewidth]{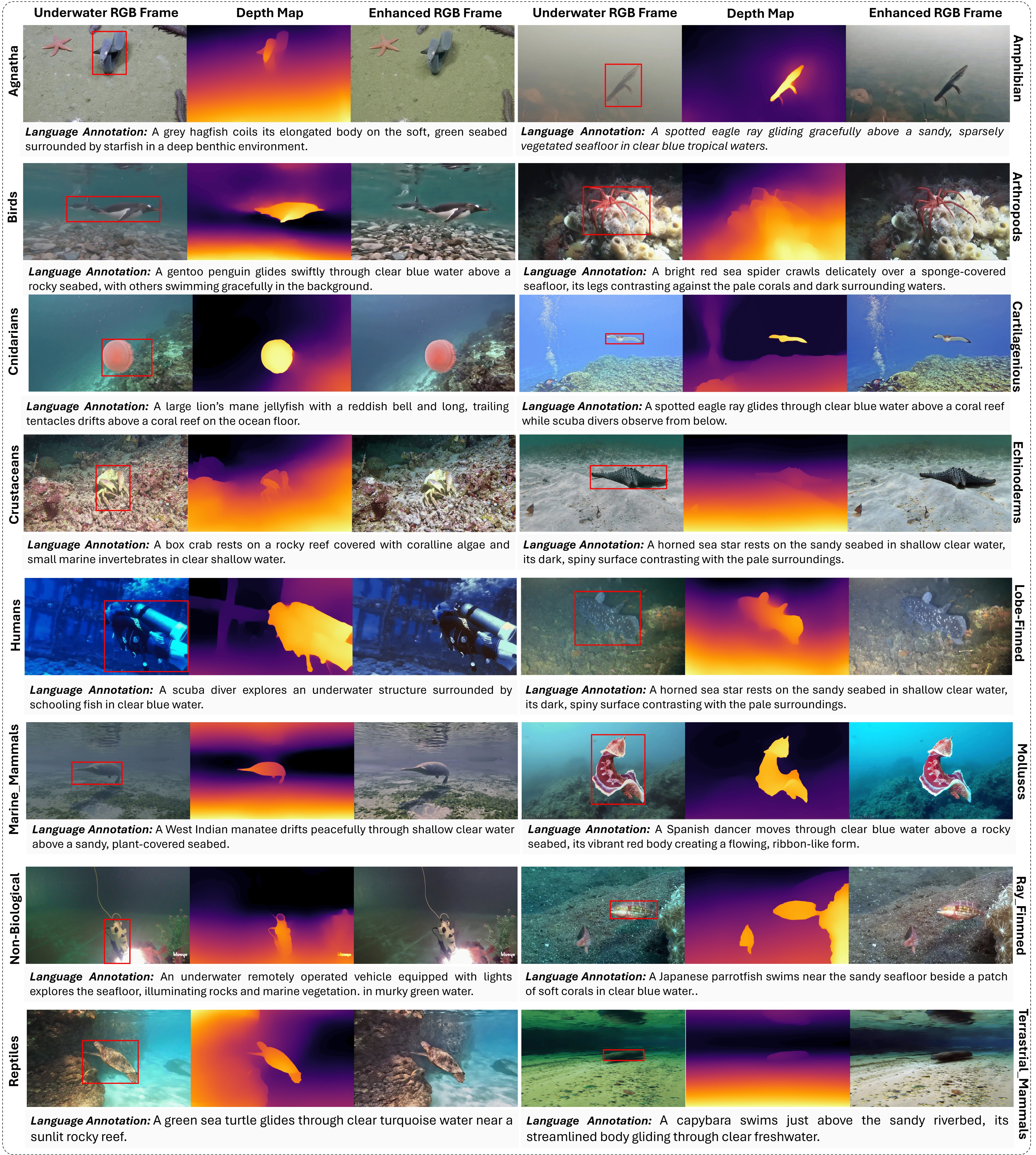}
\caption{Visualization of the \textbf{16 phylum-level classes} in the MUOT-3M dataset, showcasing examples from each category alongside their multimodal representations. 
Each triplet illustrates the RGB frame with bounding box, corresponding depth map, and enhanced image, demonstrating the dataset’s diversity across marine environments and visual conditions. 
The language annotations describe scene context and species type and behavior, emphasizing the ecological and multimodal richness of MUOT-3M Dataset.}

\label{Classes}
\end{figure*}

\subsection{MUOT-3M Diversity}
\begin{figure*}[htbp]
    \centering
    \includegraphics[width=0.93\textwidth]{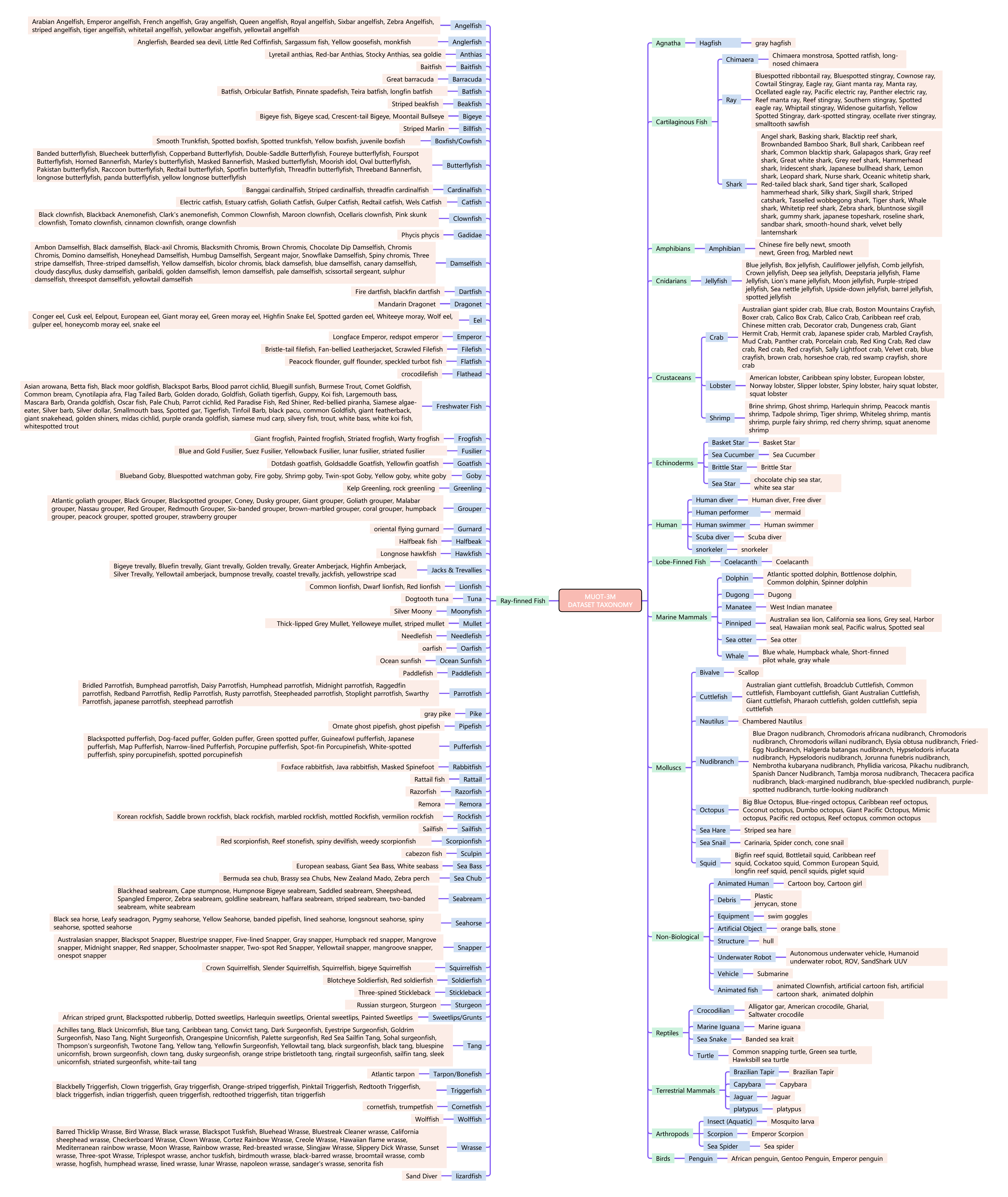}
    \caption{Hierarchical taxonomy of the \textbf{MUOT-3M Dataset}. 
    The diagram illustrates the multi-level organization of MUOT-3M, comprising 16 phylum, 124 families, and 677 fine-grained species. 
    Phylum nodes are shown in green, Family nodes in blue, and fine-grained classes in orange, representing the taxonomic depth of the dataset.}
    \label{fig:muot3m_taxonomy}
\end{figure*}

The MUOT-3M dataset exhibits extensive biological and environmental diversity, capturing a wide range of species, habitats, and modalities representative of real underwater conditions. As illustrated in Fig.~\ref{Classes}, MUOT-3M spans 16 major phylums: Ray-Finned Fish, Cartilaginous Fish, Human, Reptiles, Molluscs, Marine Mammals, Crustaceans, Cnidarians, Non-Biological, Birds, Echinoderms, Terrestrial Mammals, Agnatha, Arthropods, Amphibians, and Lobe-Finned Fish. These categories encompass over 124 families and 677 fine-grained species, covering a wide morphological spectrum from rigid-bodied vertebrates to deformable invertebrates and transparent organisms as shown in Fig.\ref{fig:muot3m_taxonomy}.
    
The dataset integrates recordings from a broad range of marine habitats, including coral reefs, coastal shallows, open-ocean environments, deep-sea regions, aquariums and artificial underwater structures. This ensures coverage of varied illumination conditions, water clarity levels, and background textures, reflecting the diverse operational contexts of underwater imaging.

MUOT-3M is constructed as a multimodal benchmark, providing synchronized RGB with bounding box, depth, and enhanced image for each sequence. These modalities jointly capture appearance and geometry, facilitating robust tracking under underwater-specific degradations such as scattering, color attenuation, and refraction. All modalities were standardized through a unified preprocessing pipeline to maintain spatial and temporal consistency across sequences.

In addition to biological and modal diversity, MUOT-3M includes a wide range of recording configurations, encompassing stationary cameras, diver-held devices, and autonomous underwater vehicles. The inclusion of both controlled and in-the-wild sequences ensures that the dataset captures the variability in motion dynamics, lighting, and visual complexity encountered in real-world underwater tracking tasks.

Through its broad taxonomic coverage, multimodal representation, and environmental heterogeneity, MUOT-3M establishes a comprehensive and ecologically diverse benchmark for evaluating generalizable multimodal tracking in underwater scenarios.

\subsection{Dataset Total Cost}
The development of the proposed underwater tracking dataset involved an estimated cost of approximately \textbf{10,000 USD}. This cost primarily reflects the extensive effort needed to \textit{validate and refine} the annotations across the full collection of \textbf{3 million underwater frames}. The initial annotation was carried out by trained PhD students using a semi-automated labeling workflow. However, due to the challenging visual characteristics of underwater environments, such as turbidity, low contrast, color distortion, occlusions, and non-rigid object motion, high-quality ground truth could only be ensured through \textit{expert verification}. Therefore, a significant portion of the expenditure was dedicated to a specialized team of experts, who reviewed, corrected, and approved the final annotations to guaranty reliability and consistency for benchmarking and long-term research use.

\section{Additional Ablation Studies}
\label{ablation}
\subsection{Impact of Depth Estimation Methods}
Table~\ref{table_depth} analyzes the impact of different depth estimation backbones used in the multimodal teacher network on the performance of the Unimodal Student Tracker (UMS).
We replace the default MiDaS depth estimator with Depth Anything \cite{depthanything} and DAC \cite{Guo2025DepthAnyCamera}, while keeping all other components fixed.
As shown in Table~\ref{table_depth}, the UMS trained with MiDaS-derived pseudo-depth achieves the highest success rates on both MUOT-3M and WebUOT-1M.
This superiority can be attributed to MiDaS’s robust domain generalization and its ability to capture consistent global structure across diverse underwater scenes.
In contrast, Depth Anything and DAC slightly underperform, likely due to their weaker robustness to underwater color degradation and turbidity.
Overall, these results confirm that reliable pseudo-depth cues play a critical role in improving geometric–photometric alignment during multimodal pretraining and subsequent unimodal distillation.

\subsection{Impact of Enhanced Underwater Methods Ablation}
Table~\ref{table_enhancement} presents an ablation study on the impact of different underwater image enhancement methods employed within the multimodal teacher network.
We evaluate three representative models: U-Transformer~\cite{peng2023u}, PixMamba~\cite{lin2024pixmamba}, and UVEB~\cite{xie2024uveb}, while keeping the remaining modules of MUTrack unchanged.
Among them, the U-Transformer yields the best results on both MUOT-3M and WebUOT-1M, achieving $66.58\%$ and $67.10\%$ success rates, respectively.
This improvement stems from its transformer-based global modeling, which better captures long-range dependencies and effectively restores fine structural details and color constancy under severe turbidity.
PixMamba and UVEB provide competitive yet slightly lower performance, likely due to their limited receptive fields and reduced capability to generalize across varying underwater lighting and scattering conditions.
These results highlight that a strong enhancement backbone, capable of producing geometrically and photometrically stable representations is essential for effective multimodal supervision and knowledge transfer in the proposed MUTrack framework.

\begin{table}[t!]
\resizebox{\linewidth}{!}{%
\centering
\begin{tabular}{l|c c c|c cc|}
\hline
Unimodal &  \multicolumn{3}{c|}{Depth Methods} &MUOT-3M &WebUOT-1M \\
Student (UMS)&MiDaS&Depth Anything&DAC&(success Rate)&(success Rate)\\
\hline
MUTrack-MMT&$\checkmark$ &  $\times$&$\times$ &\textbf{66.58}&\textbf{67.10}\\
MUTrack-MMT&$\times$ &  $\checkmark$&$\times$ &66.12&66.96\\
MUTrack-MMT&$\times$ &  $\times$&$\checkmark$ &66.40&67.06\\
\hline
\end{tabular}
}
\caption{Influence of depth estimation methods on MUTrack-UMS.
The multimodal teacher is fixed, and MUTrack student-only results are reported.
UMS stands for “Unimodal Student”}
\label{table_depth}
\end{table}

\begin{table}[t!]
\resizebox{\linewidth}{!}{%
\centering
\begin{tabular}{l|c c c|c cc|}
\hline
Unimodal &  \multicolumn{3}{c|}{Enhancement Methods} &MUOT-3M &WebUOT-1M \\
Student (UMS)&UTransformer&PixMamba&UVEB&(success Rate)&(success Rate)\\
\hline
MUTrack-MMT&$\checkmark$ &  $\times$&$\times$ &\textbf{66.58}&\textbf{67.10}\\
MUTrack-MMT&$\times$ &  $\checkmark$&$\times$ &66.50&66.93\\
MUTrack-MMT&$\times$ &  $\times$&$\checkmark$ &66.52&67.02\\
\hline
\end{tabular}
}
\caption{Influence of underwater image enhancement methods on MUTrack-UMS.
The multimodal teacher is fixed, and MUTrack student-only results are reported.
UMS stands for “Unimodal Student”}
\label{table_enhancement}
\end{table}

\section{MUOT-3M Attributes}
\label{attributes}
The MUOT-3M dataset captures a diverse range of tracking challenges, covering both generic visual factors common in general object tracking and underwater-specific factors unique to aquatic environments. 
Generic attributes describe conditions affecting target appearance and motion that are widely applicable across visual tracking tasks, 
while underwater-specific attributes capture the optical, environmental, and sensor-based effects encountered in marine scenes. 
Together, these attributes provide a comprehensive characterization of tracking difficulty and visual variability in MUOT-3M.

\subsection{Generic Attributes}

\begin{figure*}[htp]
\centering
\includegraphics[width=0.85\textwidth]{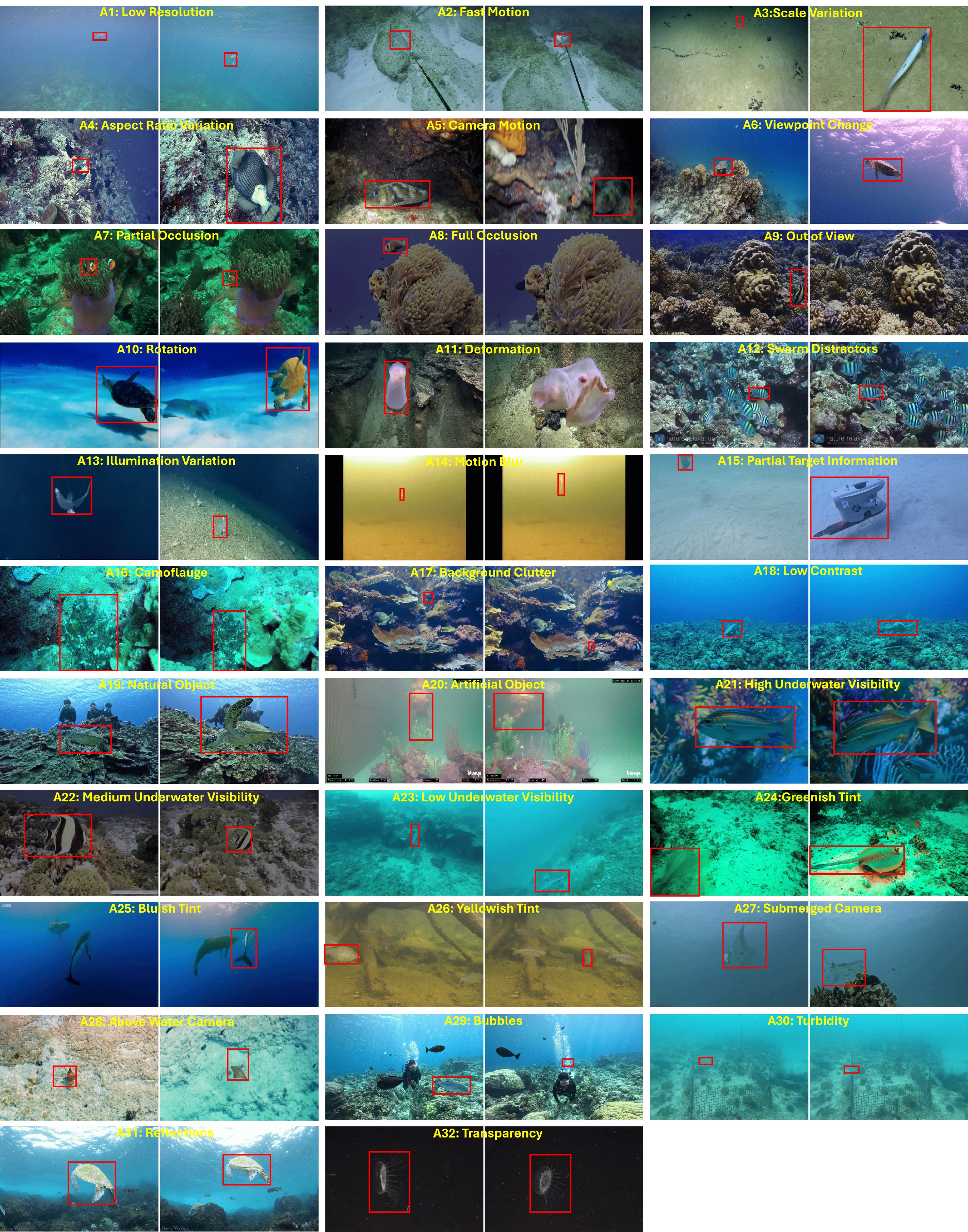}
\caption{Visualization of the 32 tracking attributes defined in the MUOT-3M dataset. 
Each attribute represents a distinct visual or environmental challenge, covering both generic and underwater-specific conditions. These attributes collectively characterize the full spectrum of appearance and motion variability in underwater tracking scenarios}

\label{Attributes}
\end{figure*}

\begin{enumerate}
    \item \textbf{Low Resolution:} Target is very small in the frame or video resolution is poor, reducing fine detail as shown in Fig. \ref{Attributes}-A1.
    \item \textbf{Fast Motion:} Target moves rapidly across consecutive frames, causing large positional shifts as shown in Fig. \ref{Attributes}-A2.
    \item \textbf{Scale Variation:} Target size changes significantly due to distance or pose as shown in Fig. \ref{Attributes}-A3.
    \item \textbf{Aspect Ratio Variation:} Target bounding box aspect ratio changes notably due to orientation or distortion as shown in Fig. \ref{Attributes}-A4.
    \item \textbf{Camera Motion:} Camera exhibits strong motion or shaking, affecting stability as shown in Fig. \ref{Attributes}-A5.
    \item \textbf{Viewpoint Change:} Target appearance alters considerably due to viewing angle change as shown in Fig. \ref{Attributes}-A6.
    \item \textbf{Partial Occlusion:} Target is partially blocked by another object in the scene as shown in Fig. \ref{Attributes}-A7.
    \item \textbf{Full Occlusion:} Target is completely blocked from view for one or more frames as shown in Fig. \ref{Attributes}-A8.
    \item \textbf{Out of View:} Target leaves the visible video frame entirely as shown in Fig. \ref{Attributes}-A9.
    \item \textbf{Rotation:} Target rotates in-plane or out-of-plane, altering its visible features as shown in Fig. \ref{Attributes}-A10.
    \item \textbf{Deformation:} Target changes shape non-rigidly (e.g., bending, stretching, flexing) as shown in Fig. \ref{Attributes}-A11.
    \item \textbf{Similar/Swarm Distractors:} Other similar-looking objects appear nearby, confusing the tracker as shown in Fig. \ref{Attributes}-A12.
    \item \textbf{Illumination Variation:} Lighting conditions on the target change due to natural or artificial sources as shown in Fig. \ref{Attributes}-A13.
    \item \textbf{Motion Blur:} Target region is blurred due to motion of the target or the camera as shown in Fig. \ref{Attributes}-A14.
    \item \textbf{Partial Target Information:} Only part of the target is visible in initial frames, limiting initialization as shown in Fig. \ref{Attributes}-A15.
    \item \textbf{Camouflage:} Target visually blends into the background, reducing distinguishability as shown in Fig. \ref{Attributes}-A16.
    \item \textbf{Background Clutter:} Scene has complex or textured background elements that distract tracking as shown in Fig. \ref{Attributes}-A17.
    \item \textbf{Low Contrast:} Target has very similar color or brightness to the background as shown in Fig. \ref{Attributes}-A18.
    \item \textbf{Natural Object:} Target belongs to natural categories like fish, marine animals and livinf things as shown in Fig. \ref{Attributes}-A19.
    \item \textbf{Artificial Object:} Target is an artificial object such as tools, robots, or man-made equipment as shown in Fig. \ref{Attributes}-A20.
\end{enumerate}
\subsection{Underwater-Specific Attributes}

\begin{enumerate}
    \setcounter{enumi}{20} 
    \item \textbf{High Underwater Visibility:} Water is clear with minimal scattering, target easily visible as shown in Fig. \ref{Attributes}-A21.
    \item \textbf{Medium Underwater Visibility:} Water has moderate clarity with some scattering or turbidity as shown in Fig. \ref{Attributes}-A22.
    \item \textbf{Low Underwater Visibility:} Water is murky, target visibility is poor as shown in Fig. \ref{Attributes}-A23.
    \item \textbf{Greenish Tint:} Green water, affecting brightness and contrast as shown in Fig. \ref{Attributes}-A24.
    \item \textbf{Bluish Tint:} Water environment appears light blue, altering illumination as shown in Fig. \ref{Attributes}-A25.
    \item \textbf{Yellowish Tint:} Water environment appears light yellow, often indicating turbidity as shown in Fig. \ref{Attributes}-A26.
    \item \textbf{Submerged Camera:} Video is recorded directly underwater as shown in Fig. \ref{Attributes}-A27.
    \item \textbf{Above-water Camera:} Video is recorded from outside the water, often distorted by refraction as shown in Fig. \ref{Attributes}-A28.
    \item \textbf{Bubbles:} Air bubbles appear in the scene, acting as distractors or occlusions as shown in Fig. \ref{Attributes}-A29.
    \item \textbf{Turbidity:} Suspended particles reduce water clarity, creating noise as shown in Fig. \ref{Attributes}-A30.
    \item \textbf{Reflections:} Reflections from surfaces (water, glass, metal) distort or duplicate the target as shown in Fig. \ref{Attributes}-A31.
    \item \textbf{Transparency:} Target is semi-transparent (e.g., jellyfish), making boundaries hard to detect as shown in Fig. \ref{Attributes}-A32.
\end{enumerate}

\section{Attribute-based Performance} 
\label{attribute_performance}
We have also conducted the attribute-wise performance comparison of our proposed MUTRack with SOTA trackers trained in protocol II on 32 underwater tracking attributes of the MUOT-3M dataset, as shown in the supplementary material. 
Across all 32 tracking attributes, our proposed MUTrack has consistently outperformed SOTA trackers, especially in terms of underwater-specific attributes such as water color variation, camouflage, and visibility conditions.

Across all 32 attribute subsets of MUOT-3M, MUTrack-UMS exhibits strong and consistent performance, achieving top AUC values in the majority of conditions while maintaining competitive accuracy where it is not the outright leader. It delivers the highest success under rotation (70.2\%), scale variation (67.2\%), aspect ratio variation (66.9\%), and medium visibility (66.8\%), highlighting its effective handling of geometric deformations and variable underwater clarity. Furthermore, MUTrack-UMS excels under underwater-specific challenges such as turbidity (53.5\%), transparency (53.5\%), and swarm distractors (61.2\%), confirming its resilience to scattering, camouflage, and visual ambiguity in dense marine scenes. Even in cases where other methods slightly surpass it, MUTrack’s results remain close and up to par, reflecting its general robustness and reliable generalization across visual domains.

DUTrack demonstrates complementary strengths, outperforming others under illumination variation (68.3\%), motion blur (64.1\%), and partial target information (63.7\%), owing to its enhanced multimodal fusion and adaptive pre-processing strategy. ATCTrack also achieves leading scores in full occlusion (53.4\%), partial occlusion (60.2\%), and artificial object tracking (72.6\%), revealing strong target reinitialization and feature retention in structured object categories. Meanwhile, ARTrackV2 and SuperSBT perform competitively in balanced attributes such as background clutter, viewpoint change, and reflections, offering steady performance despite lacking multimodal input. These marginal differences across attributes underline the diversity and fairness of MUOT-3M as a benchmark, where each tracker’s architecture finds its unique operational niche.

Across all methods, performance trends reveal that tracking accuracy peaks in high-visibility (72.9\%) and submerged-camera (67.8\%) sequences, where optical conditions are stable, and degrades notably in bubbles, camouflage, and above-water refraction scenarios, which introduce heavy noise and visual distortion. Overall, the attribute-wise comparison demonstrates that while MUTrack-UMS provides the most balanced and reliable tracking across both generic and underwater-specific conditions, DUTrack and ATCTrack show situational advantages in specific lighting and occlusion scenarios. Together, these results validate MUTrack’s competitive and consistent behavior, showing that even where minor performance gaps exist, it remains fully on par with or above state-of-the-art counterparts.
Please see Figs below.


\begin{figure*}[t]
\centering
\includegraphics[width=\textwidth]{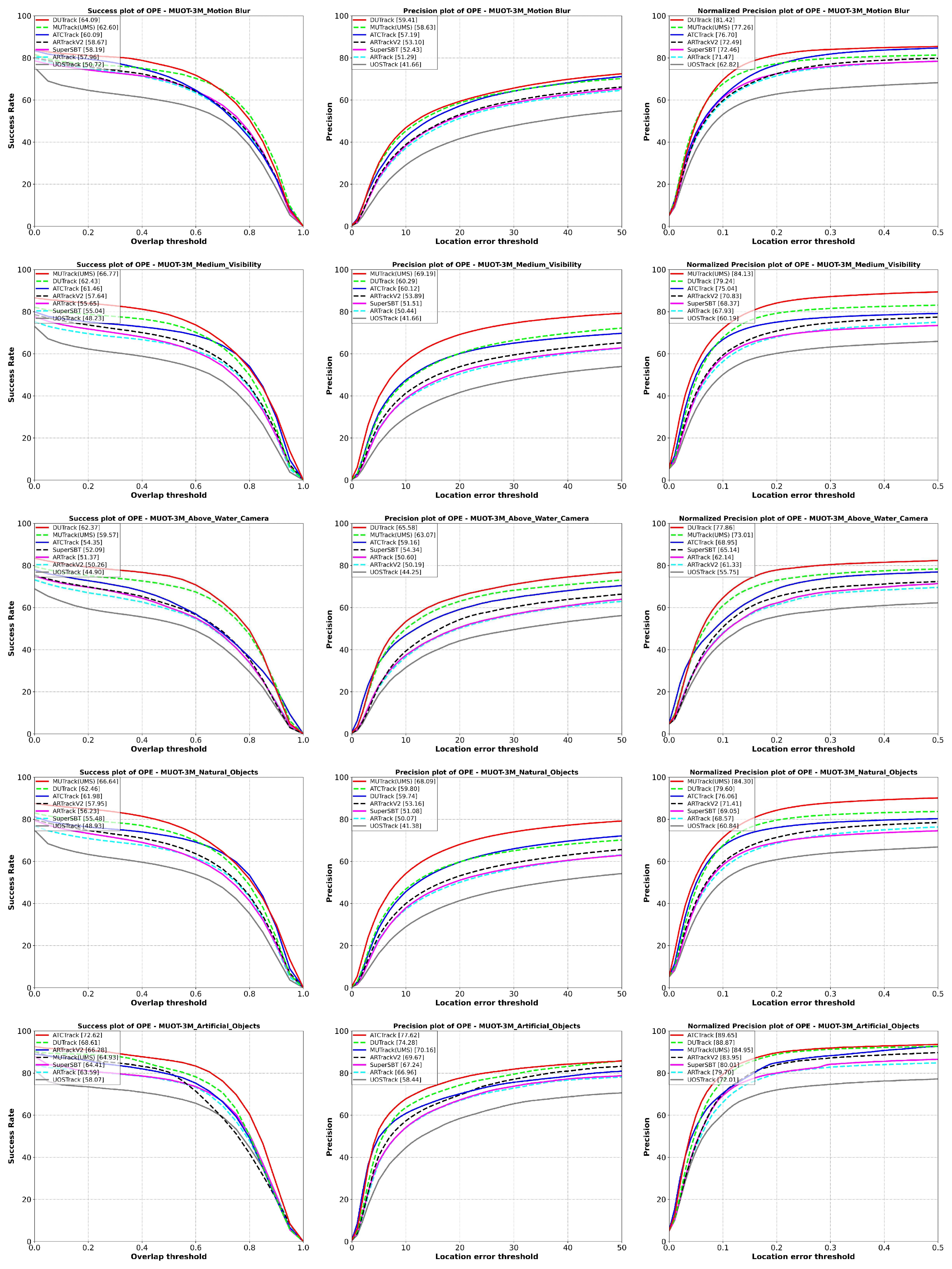}
\caption{Attribute-wise tracking performance (1--5)}
\end{figure*}
\clearpage

\begin{figure*}[t]
\centering
\includegraphics[width=\textwidth]{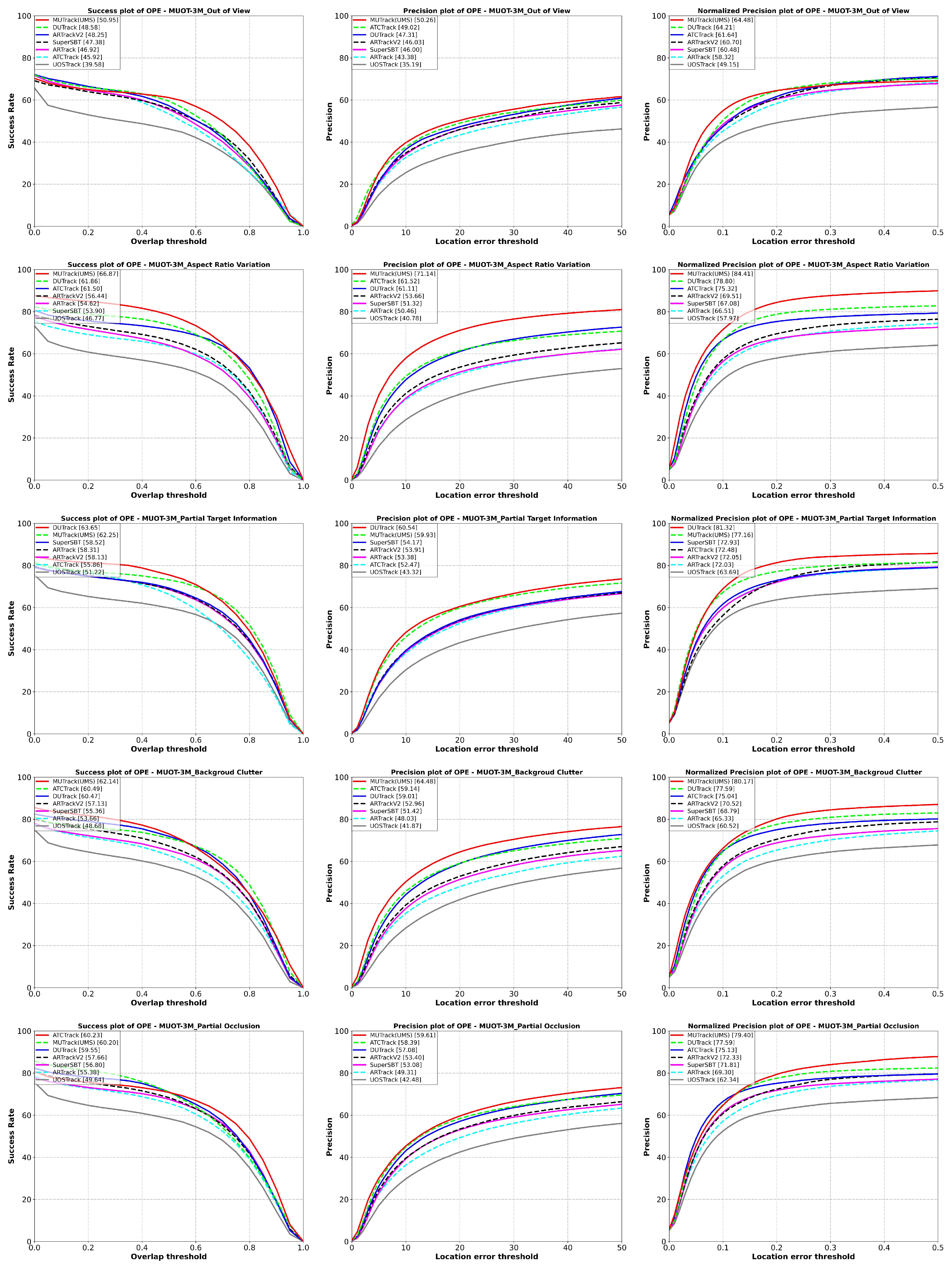}
\caption{Attribute-wise tracking performance (6--10)}
\end{figure*}
\clearpage

\begin{figure*}[t]
\centering
\includegraphics[width=\textwidth]{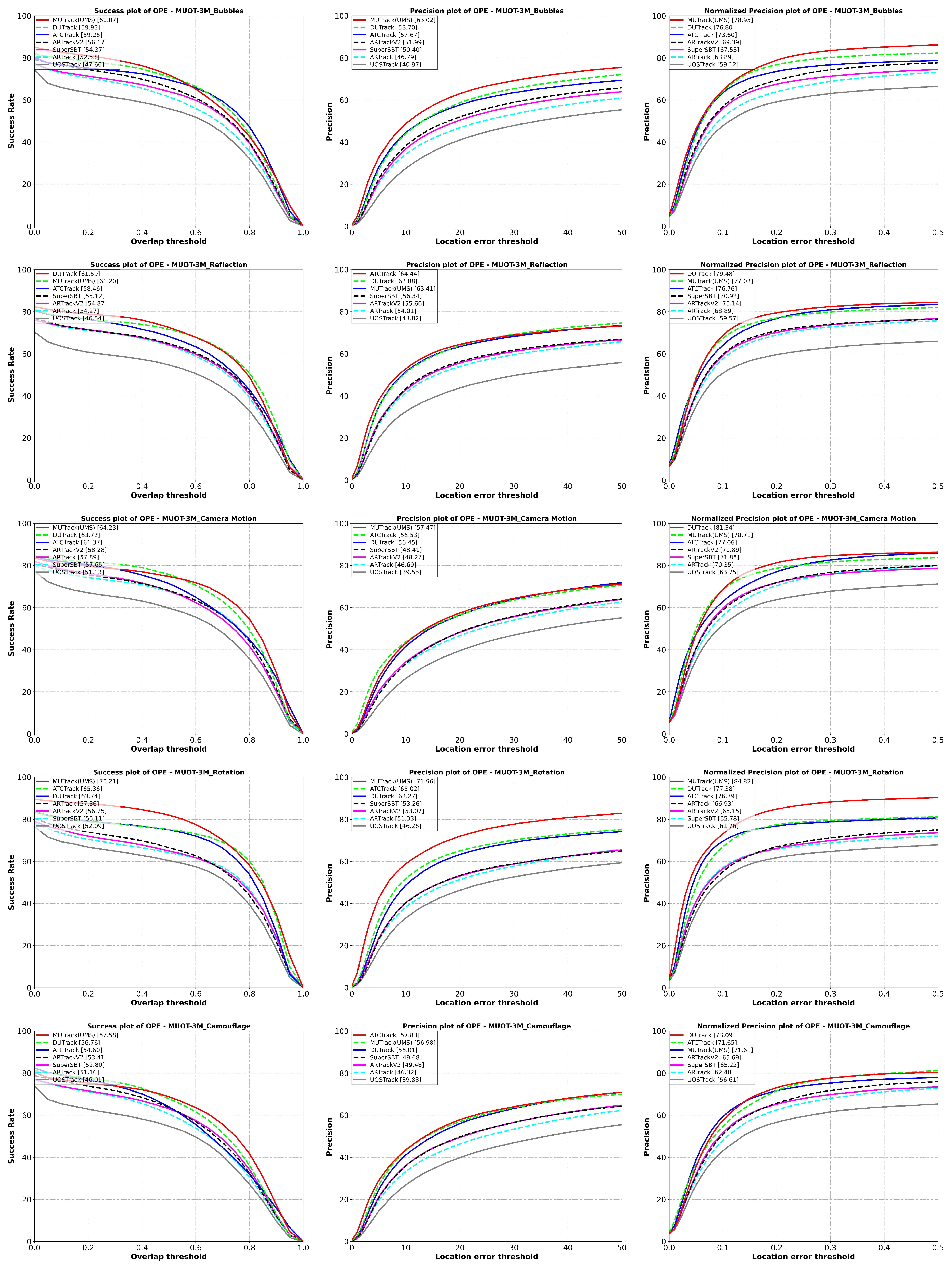}
\caption{Attribute-wise tracking performance (11--15)}
\end{figure*}
\clearpage

\begin{figure*}[t]
\centering
\includegraphics[width=\textwidth]{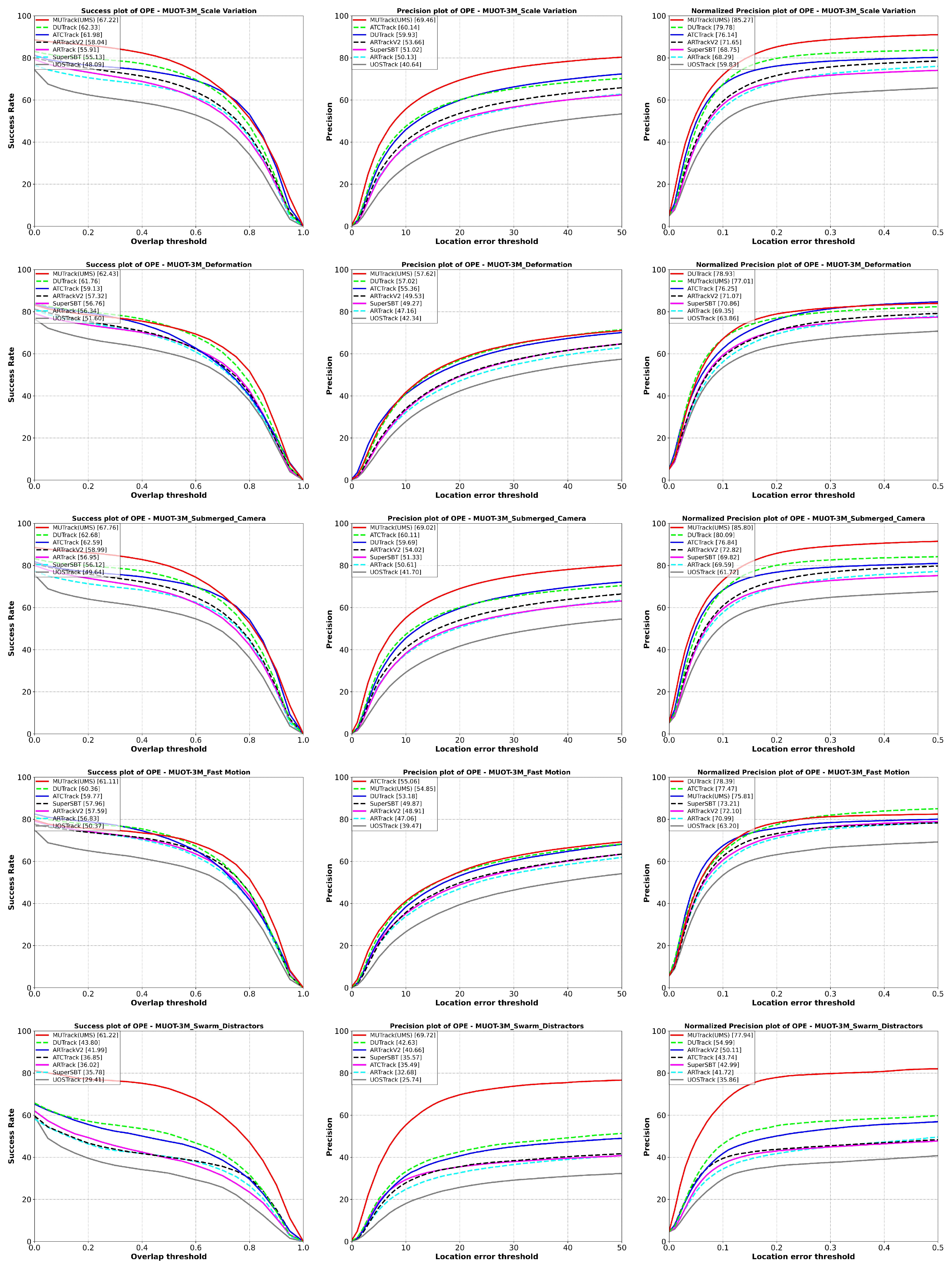}
\caption{Attribute-wise tracking performance (16--20)}
\end{figure*}
\clearpage

\begin{figure*}[t]
\centering
\includegraphics[width=\textwidth]{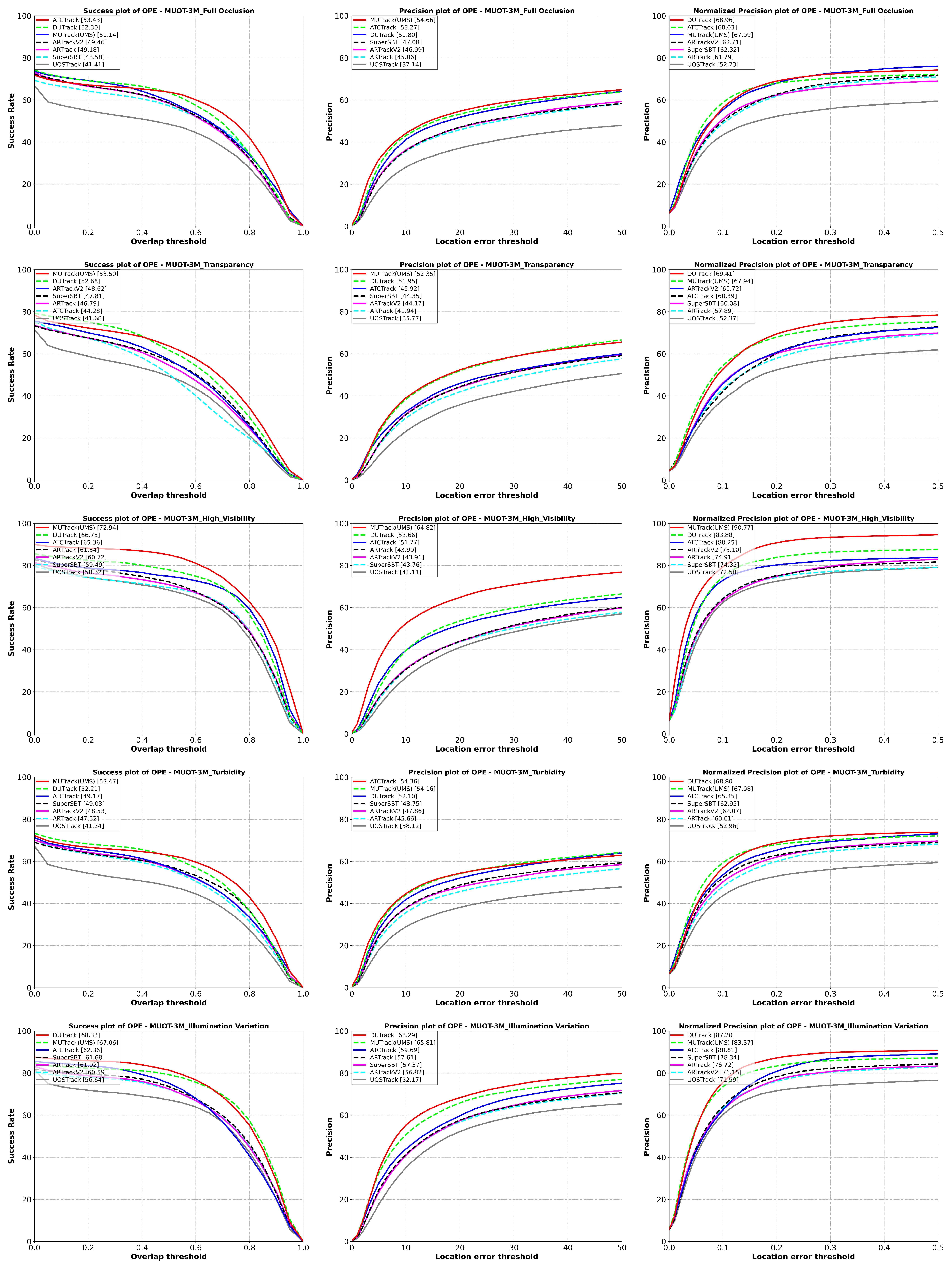}
\caption{Attribute-wise tracking performance (21--25)}
\end{figure*}
\clearpage

\begin{figure*}[t]
\centering
\includegraphics[width=\textwidth]{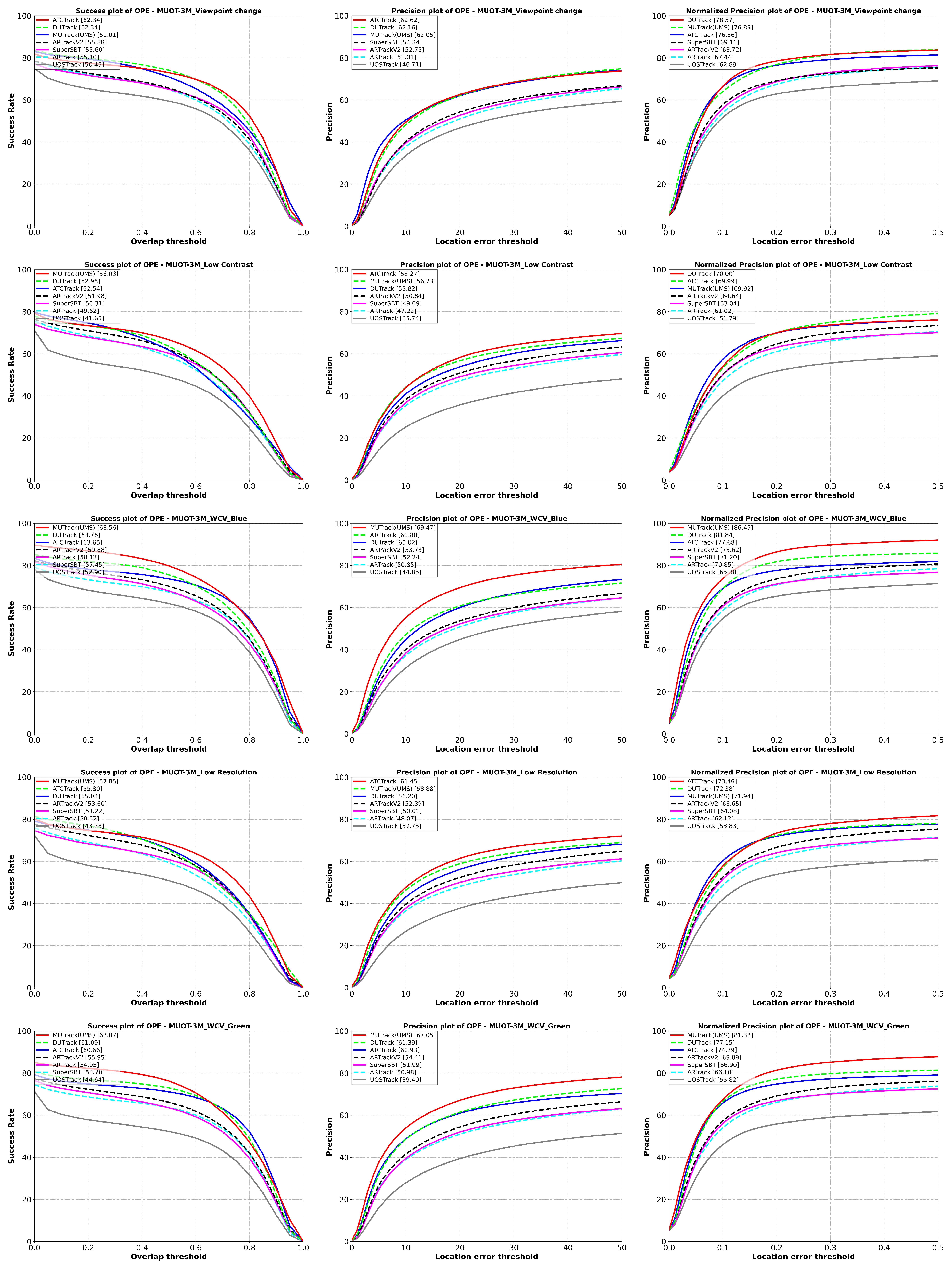}
\caption{Attribute-wise tracking performance (26--30)}
\end{figure*}
\clearpage

\begin{figure*}[t]
\centering
\includegraphics[width=\textwidth]{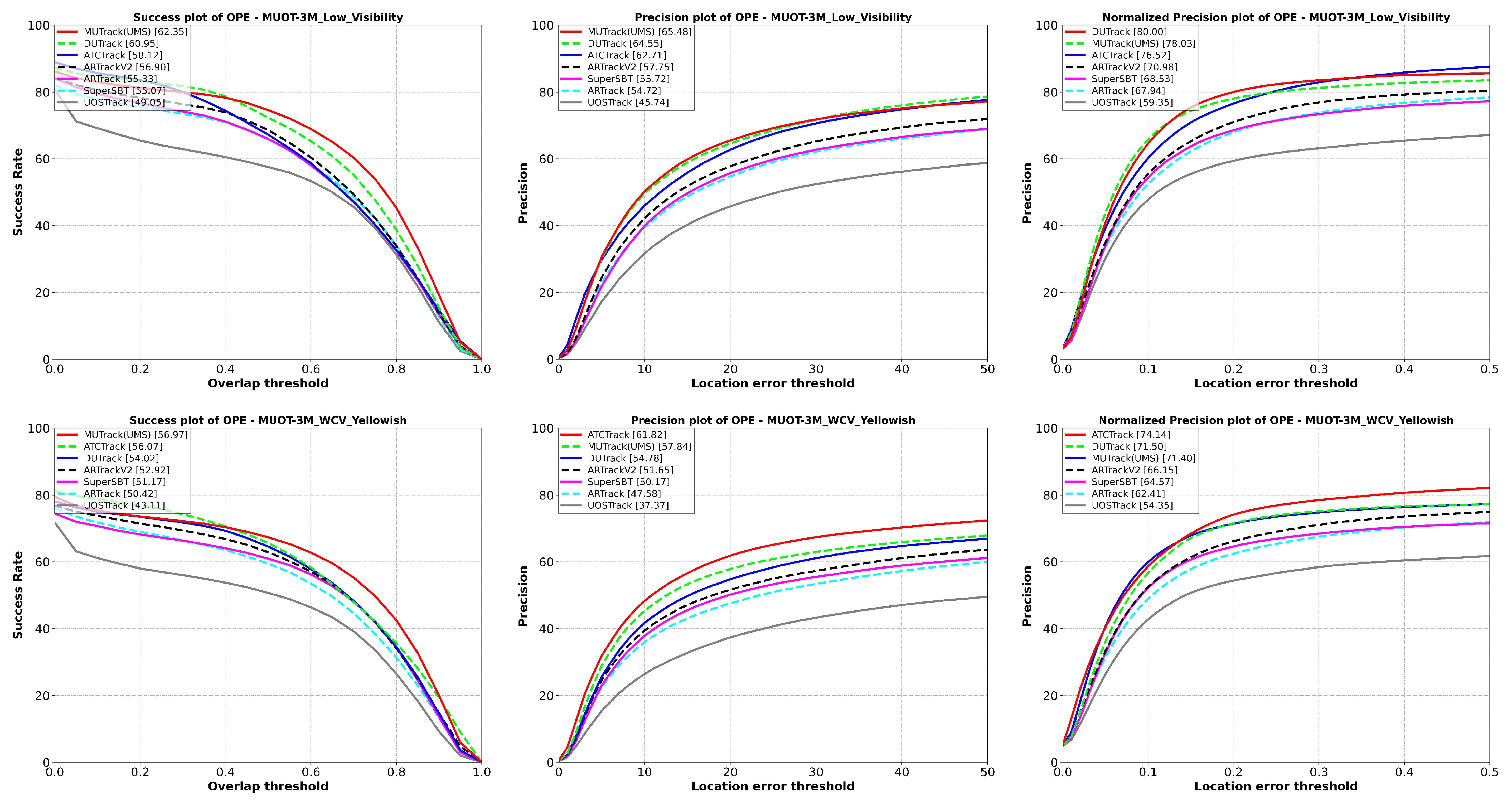}
\caption{Attribute-wise tracking performance (31--32)}
\end{figure*}